\documentclass[journal]{IEEEtran}

\usepackage {
	amsmath,
	graphicx,
	mathrsfs,
	graphicx,
	multirow,
	multicol,
	color,
	array,
	amssymb,
	bm,
	tabularx,
	xcolor,
}

\usepackage[ruled]{algorithm2e}

\usepackage[
breaklinks=true,
bookmarks=false,
colorlinks,
linkcolor=blue,
anchorcolor=red,
citecolor=red]{hyperref}

\usepackage[noadjust]{cite}

\usepackage[misc]{ifsym}
\usepackage[export]{adjustbox}

\usepackage{pifont}
\ifCLASSINFOpdf
\else
\fi
\begin{document}
	%
	\title{Content-aware Scalable Deep Compressed Sensing}
	%
	%
	%
	
	\author{Bin~Chen,
		Jian~Zhang,~\IEEEmembership{Member,~IEEE}\vspace{-30pt}

		\thanks{Manuscript received August 9, 2021; revised April 12, 2022; accepted July 12, 2022. This work was supported by National Natural Science Foundation of China (61902009).}
				
		\thanks{Bin Chen and Jian Zhang are with the School of Electronic and Computer Engineering, Peking University Shenzhen Graduate School, Shenzhen 518055, China. J. Zhang is also with the Peng Cheng Laboratory, Shenzhen,
China. (e-mail: chenbin74851@126.com; zhangjian.sz@pku.edu.cn).}

		}
	
	%
	%

	\markboth{Journal of \LaTeX\ Class Files,~Vol.~xx, No.~xx, ~2022}%
	{Shell \MakeLowercase{\textit{et al.}}: Bare Demo of IEEEtran.cls for IEEE Journals}
	%



	\maketitle
	\begin{abstract}
		To more efficiently address image compressed sensing (CS) problems, we present a novel content-aware scalable network dubbed CASNet which collectively achieves adaptive sampling rate allocation, fine granular scalability and high-quality reconstruction. We first adopt a data-driven saliency detector to evaluate the importances of different image regions and propose a saliency-based block ratio aggregation (BRA) strategy for sampling rate allocation. A unified learnable generating matrix is then developed to produce sampling matrix of any CS ratio with an ordered structure. Being equipped with the optimization-inspired recovery subnet guided by saliency information and a multi-block training scheme preventing blocking artifacts, CASNet jointly reconstructs the image blocks sampled at various sampling rates with one single model. To accelerate training convergence and improve network robustness, we propose an SVD-based initialization scheme and a random transformation enhancement (RTE) strategy, which are extensible without introducing extra parameters. All the CASNet components can be combined and learned end-to-end. We further provide a four-stage implementation for evaluation and practical deployments. Experiments demonstrate that CASNet outperforms other CS networks by a large margin, validating the collaboration and mutual supports among its components and strategies. Codes are available at \url{https://github.com/Guaishou74851/CASNet}.
  \end{abstract}
	
	\begin{IEEEkeywords}
		Compressed sensing, image restoration, content-aware sampling, model scalability, deep unfolding network.
	\end{IEEEkeywords}

	%
	\IEEEpeerreviewmaketitle
	
	\begin{figure*}[t]
		\centering
		\vspace{-16pt}
		\includegraphics[width=0.98\textwidth]{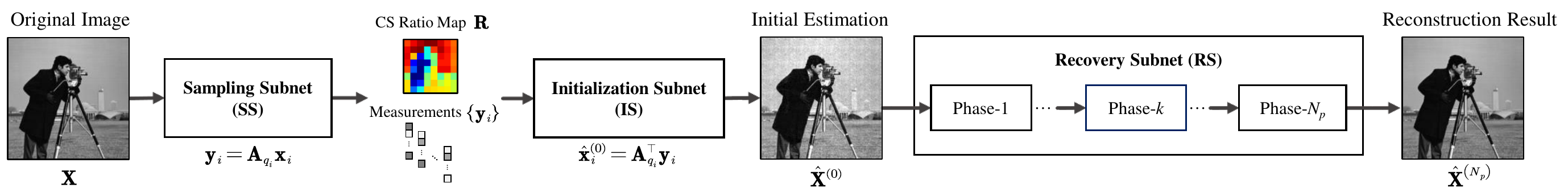}
		\vspace{-10pt}
		\caption{Illustration of our CASNet framework, which consists of a sampling subnet (SS, Fig. \ref{fig: ss}), an initialization subnet (IS, Fig. \ref{fig: is}), and a recovery subnet (RS, Fig. \ref{fig: rs}). In block-based CS scheme, the original image $\mathbf{X}$ is divided into $l$ non-overlapped $B\times B$ blocks $\{\mathbf{x}_i\}$. In SS, they are reshaped to be $N\times 1$ ($N=B^2$) and linearly sampled one-by-one to obtain block measurements $\{\mathbf{y}_i|\mathbf{y}_i\in\mathbb{R}^{q_i\times 1}, 0\le q_i \le N\}$, which are then initialzed by IS to get a joint recovered estimation $\mathbf{X}$ by RS. Under this framework, CASNet can achieve content-aware sampling and fine granular scalability with only one single network.}
		\label{fig: arch}
		\vspace{-14pt}
	\end{figure*}
	
	\vspace{-4pt}
	\section{Introduction}
	\IEEEPARstart{C}{ompressed} sensing (CS) is a novel paradigm that requires much fewer measurements than the Nyquist sampling for signal acquisition and restoration \cite{CS, CSintro}. For the signal $ \mathbf{x}\in \mathbb{R}^N $, it conducts the sampling process $ \mathbf{y}=\mathbf{\Phi x} $ to obtain the measurements $ \mathbf{y}\in \mathbb{R}^M $, where $ \mathbf{\Phi} \in \mathbb{R}^{M \times N} $ with $ M\ll N $ is a given sampling matrix, and the CS ratio (or sampling rate) is defined as $r= {M}/{N} $. Since it is hardware-friendly and has great potentials of improving sampling speed with high recovery accuracy, many applications have been developed including single-pixel imaging \cite{SinglePixelImaging1, SinglePixelImaging2}, magnetic resonance imaging (MRI) \cite{MRI1, MRI2}, sparse-view CT \cite{szczykutowicz2010dual}, etc. In this work, we focus on the typical block-based (or block-diagonal) image CS problem \cite{Block1, chun2017compressed, chun2020uniform} that divides the high-dimensional natural image into non-overlapped $B\times B$ blocks and obtains measurements block-by-block with a small fixed sampling matrix for the subsequent reconstruction.
	
	Recovering $ \mathbf{x} $ from the acquired $\mathbf{y}$ is to solve an ill-posed under-determined inverse system. Many model-driven methods focus on exploiting structural prior with theoretical guarantees, such as sparse representation \cite{zhangL0min, zhao2016video, zhao2016reducing, elad2010sparse, nam2013cosparse}, low-rank \cite{LowRank1, LowRank2, LowRank3, LowRank4, LowRank5}, etc. However, they are suffering from high computational costs and rely on extensive fine-tuning and empirical results.
	
	Recently, the development of deep convolutional neural network (CNN) greatly improves recovery accuracy and speed \cite{ravishankar2019image}. By adopting end-to-end pipelines, \cite{DL1, DL2_ReconNet, DL3_CSNet} can quickly perform recoveries but may leave undesired effects \cite{BlackBox}. Motivated by the classic structure-texture decomposition paradigm, Sun \textit{et al.} \cite{sun2020dual} propose a dual-path attention CS network. Deep unfolding methods \cite{DeepUnfolding1, DeepUnfolding2, DeepUnfolding3, ADMM-Net, AMP-inspired, ISTA-Net, chun2020momentum} map restoration algorithms into network architectures and achieve balances between speed and interpretation. Based on traditional ISTA \cite{ISTA} and AMP \cite{AMP}, J. Zhang \textit{et al.} \cite{OPINE-Net} and Z. Zhang \textit{et al.} \cite{AMP-Net} respectively adopt learnable sampling matrices and propose more powerful CNNs with inter-block relationship exploitation and deep structural insights. Chun \textit{et al.} \cite{chun2020momentum} further introduce momentum mechanism in each unfolded iteration for recovery acceleration.
	
	In addition to the network structure and optimization algorithm design, the CS ratio allocation and model scalability have been concerned and studied. By considering the characteristics of human visual system, Yu \textit{et al.} \cite{SaliencyBased} adopt a strategy of allocating less sampling rates to non-salient blocks and more to salient ones. Zhou \textit{et al.} \cite{BCS-Net} develop a multi-channel network to obtain high restoring accuracy with a similar allocating mechanism. To balance the CS system complexity and flexibility, Shi \textit{et al.} \cite{SCSNet} propose a hierarchical CNN to achieve scalable sampling and recovery. Recently, You \textit{et al.} \cite{you2021coast} propose a robust controllable network dubbed COAST to deal with the recoveries under arbitrary sampling matrices.
	
	Although most state-of-the-art methods yield high performances, the problems they focus on are still not comprehensive enough, thus leading to many their restricted applications. In this paper, we propose a novel \textbf{C}ontent-\textbf{A}ware \textbf{S}calable\footnote{Following \cite{SCSNet, you2021coast}, we call a CS network scalable (or fine granular scalable) when it uses one set of parameters for a single network to handle multiple CS ratios (or all CS ratios in the common range of $[0,0.5]$ or $[0,1]$).} deep \textbf{Net}work dubbed \textbf{CASNet} to comprehensively solve natural image CS problems. Specifically, we explore the structural potentials of CASNet by developing the existing approaches and trying to keep and organically combine their merits. As illustrated in Fig.~\ref{fig: arch}, CASNet is composed of a sampling subnet (SS), an initialization subnet (IS) and a recovery subnet (RS). We propose a unified learnable generating matrix to produce sampling matrices and a data-driven saliency detector so that our CASNet can perform efficient and adaptive CS ratio allocations with the fine granular model scalability. The multi-phase recovery subnet can explore the inter-block relationship under the saliency information guidance. With the well-defined structure and mutual supports among different components, CASNet can be learned in a completely end-to-end manner and enjoys the merits of accurate recovery and interpretability.
	
	The main three contributions of this paper are as follows:
	
    \noindent \ding{113} (1) A content-aware scalable network dubbed CASNet is proposed to achieve block-wise CS ratio allocation and handle image CS task under any sampling rate $r\in[0,1]$ by a single network. To our knowledge, this is the first work integrating CS ratio allocation, model scalability and unfolded recovery.
    
	\noindent \ding{113} (2) We propose to adopt a lightweight CNN to adaptively detect the image saliency distribution, and design a block ratio aggregation (BRA) strategy to achieve block-wise CS ratio allocation instead of using a handcrafted detecting function adopted by previous saliency-based methods \cite{SaliencyBased, BCS-Net}.
	
	\noindent \ding{113} (3) We further provide two boosting strategies and a four-stage implementation for CASNet evaluations. Experiments show that CASNet outperforms state-of-the-art CS methods with benefiting from the inherent strong compatibility and mutual supports among its different components and strategies.
	
	\vspace{-6pt}
	\section{Related Works}
	We group existing CS approaches into \textit{optimization-} and \textit{network-based} methods. In this section, we will retrospect both and focus on the specific methods most relevant to our own.
	
	\textbf{Optimization-based Methods:} Traditional CS approaches usually recover $\mathbf{x}$ from the acquired $\mathbf{y}$ by solving the following optimization problem which is often assumed to be convex:
	\begin{equation}
		\label{traditional_optimization_problem}
		\begin{split}
		    \vspace{-4pt}
			\mathbf{\hat{x}}=\underset{\mathbf{x}}{\arg\min}\frac{1}{2}\lVert \mathbf{\Phi {x}}-\mathbf{y} \rVert _{2}^{2} +\lambda \mathcal{R}\left( \mathbf{{x}} \right) ,
			\vspace{-4pt}
		\end{split}
	\end{equation}
	here $\lambda \mathcal{R}\left( \mathbf{{x}} \right)$ is a prior term with regularization parameter $\lambda$.
	
	Many of the classic domains \cite{Block2,ADMM} and prior knowledge about transform coefficients \cite{Iter1,structure} have been exploited to reconstruct images by means of various iterative solvers (\textit{e.g.} ISTA \cite{ISTA}, ADMM \cite{ADMM} and AMP \cite{AMP}). And there are lots of methods based on image nonlocal properties \cite{Iter2, CollaborativeSparsity, Nonconvex, LowRank1} and denoiser-integrating techniques \cite{metzler2016denoising, IRCNN, rick2017one, zhao2018cream}. Furthermore, some data-driven methods are proved to be robust and effective including dictionary learning \cite{aharon2006k}, tight-frame learning \cite{cai2014data} and convolutional operator learning \cite{chun2019convolutional}. However, these methods give rise to high computational cost and are suffering from challenging prior or parameter settings and fine-tunings.
	
	\textbf{Network-based Methods:} Recently, the network-based CS methods demonstrate their promising performances. Kulkarni \textit{et al.} \cite{DL2_ReconNet} propose to learn a CNN to regress an image patch from its corresponding measurement. Shi \textit{et al.} \cite{shi2017deep} propose a framework called CSNet which avoids blocking artifacts by learning a mapping between block measurements and jointly recovered image. Sun \textit{et al.} \cite{sun2020dual} propose a dual-path attention network dubbed DPA-Net, whose structural and textural paths are bridged by a texture attention module. Lately, unfolding networks are developed to combine the merits of optimization-based methods and network-based methods. Zhang \textit{et al.} \cite{ISTA-Net} develop a so-called ISTA-Net$^+$ which works well for CS and CS-MRI tasks. J. Zhang \textit{et al.} \cite{OPINE-Net} and Z. Zhang \textit{et al.} \cite{AMP-Net} further exploit the inter-block relationship and propose the networks dubbed OPINE-Net$^+$ and AMP-Net, respectively.
	
	However, most existing network-based methods regard the CS sampling-reconstruction under different sampling rates as different tasks. They train a set of network parameters for only a specific CS ratio, and need to learn $N$ networks to support all sampling rates in $\{q/N\}_{q=1}^N$. This causes the complex and huge CS system (with storing all parameters), which is expensive for hardware implementation. By considering CS system memory cost the importance of model scalability, Shi \textit{et al.} \cite{SCSNet} propose a scalable CNN named SCSNet which adopts a hierarchical structure and a heuristic greedy method performed on an auxiliary dataset to separately learn and sort measurement bases, but this brings its training difficulty and the defect of delicacy. Inspired by the block-wise sampling rate allocation mechanism in \cite{SaliencyBased} and a block-based CS algorithm in \cite{Block1}, Zhou \textit{et al.} \cite{BCS-Net} propose a multi-channel framework dubbed BCS-Net using a channel-specific sampling network to achieve adaptive CS ratio allocation. However, the handcrafted and fixed saliency detecting method based on DCT causes its weak adaptability, and the structural inadequacy of multi-channel framework brings its inflexibility and low efficiency. Recently, You \textit{et al.} \cite{you2021coast} solve the CS problems of arbitrary-sampling matrices by a controllable network named COAST with introducing a random projection augmentation (RPA) strategy to promote training diversity, but its sampling matrices are independently generated and lack adaptability with recovery network, and it needs hundreds of sampling matrices with several pre-defined CS ratios for training, thus leading to its expensive learning and restricted performance.
	
	\vspace{-6pt}
	\section{Proposed Method}
	In this section, we first give an overview of our main ideas, elaborate on the details of CASNet framework design, then describe the integrated model, all involved parameters of which can be jointly trained end-to-end, and finally provide an implementation scheme for evaluations and deployments.
	
	\vspace{-10pt}
	\subsection{Overview of Main Ideas}
\begin{figure}
	    \vspace{-16pt}
		\hspace{-4pt}
		\includegraphics[width=0.5\textwidth]{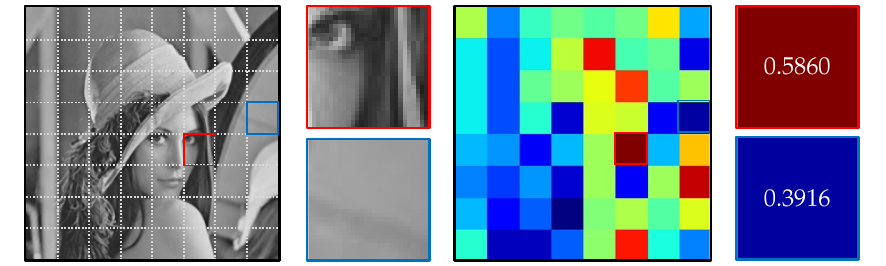}
		\vspace{-20pt}
		\caption{Illustration of an image named ``Lena" from Set11 \cite{DL2_ReconNet} with non-uniform information distribution (\textcolor{blue}{left}) and its content-aware adaptive sampling rate allocation result with keeping an average CS ratio $r=50\%$ (\textcolor{blue}{right}). We can assign higher sampling rates to the more salient blocks (\textit{e.g.} the block in red box) compared to the less salient ones (\textit{e.g.} the block in blue box).}
		\label{fig: b_comp}
		\vspace{-10pt}
	\end{figure}

	\vspace{-2pt}
	\label{main_ideas}
	\textbf{(1) Saliency-based CS ratio allocation.} Block-based CS \cite{Block1, chun2017compressed, chun2020uniform} is effective for processing high-dimensional images. In particular, we adopt an adaptive sampling rate allocation scheme that was preliminarily studied in \cite{SaliencyBased, BCS-Net} with the human perception consideration. Since image information is not always evenly distributed, one way to get restored image quality improvements is to make better CS ratio allocations by using the saliency distribution. Here we use the definition of visual saliency in \cite{yu2009hebbian}, that is, a location with low spatial correlation with its surroundings is salient. As Fig.~\ref{fig: b_comp} shows, for the given example image, the block in red box should be assigned a higher CS ratio compared to the one in blue box due to its more complex details with richer information.
	
\begin{figure}
		\includegraphics[width=0.48\textwidth]{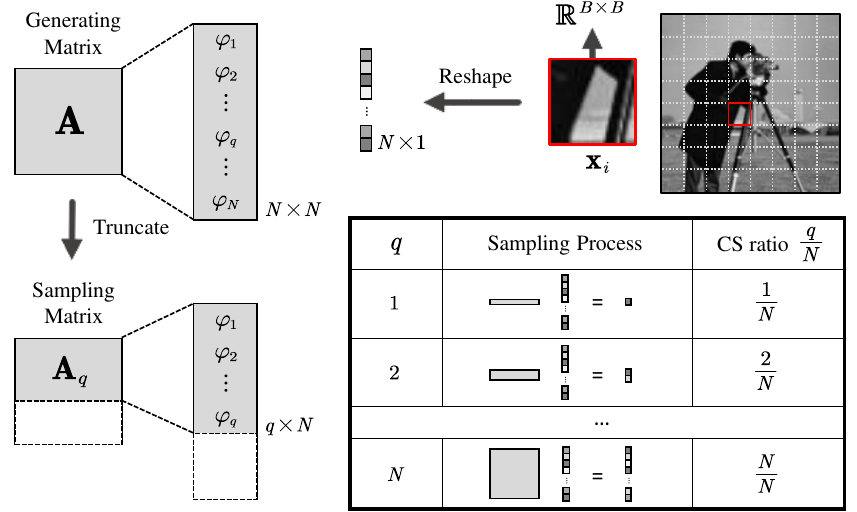}
		\vspace{-10pt}
		\caption{Illustration of our learnable generating matrix $\mathbf{A}$ with a descending base importance order for achieving fine granular model scalability.}
		\label{fig: generating_matrix}
		\vspace{-12pt}
	\end{figure}

 	\textbf{(2) Fine granular scalable sampling based on a unified learnable generating matrix.} The design of sampling matrix $\mathbf{\Phi}$, which is composed of one or more measurement bases $\{\varphi_i\}_{i=1}^M$, has been one of the main challenges in CS fields \cite{SCSNet}. Motivated by the low-rank theory and \cite{SCSNet}, we propose to obtain all sampling matrices from a unified learnable generating matrix $\mathbf{A}$ with a decreasing trend of base importance from the first row to the last, \textit{i.e.} the generating matrix is designed to generate the sampling matrix for any CS ratio and can be learned from data. As Fig.~\ref{fig: generating_matrix} illustrates, by preserving the most important $q$ measurement bases, we can obtain the sampling matrix $ \mathbf{A}_q=\mathbf{A}\left[1:q\right]$ for CS ratio $r={q}/{N}$, where $\mathbf{A}[i:j]\in\mathbb{R}^{(j-i+1)\times N}$ denotes a truncated matrix that consists of the $i$-th to the $j$-th rows of generating matrix $\mathbf{A}$. With this idea, the model scalability can be achieved with better insights into the matrix structure. Compared with most methods that need to train $N$ sampling matrices with a total memory cost of $\sum_{q=1}^{N}(qN)=[N^2(N+1)/2]\in \mathcal{O}(N^3)$ for all CS ratios $\{q/N\}_{q=1}^N$, our generating matrix takes a largely reduced storage complexity of $(N^2)\in \mathcal{O}(N^2)$.
	
	\begin{figure}[t]
		\centering
		\vspace{-16pt}
		\includegraphics[width=0.50\textwidth]{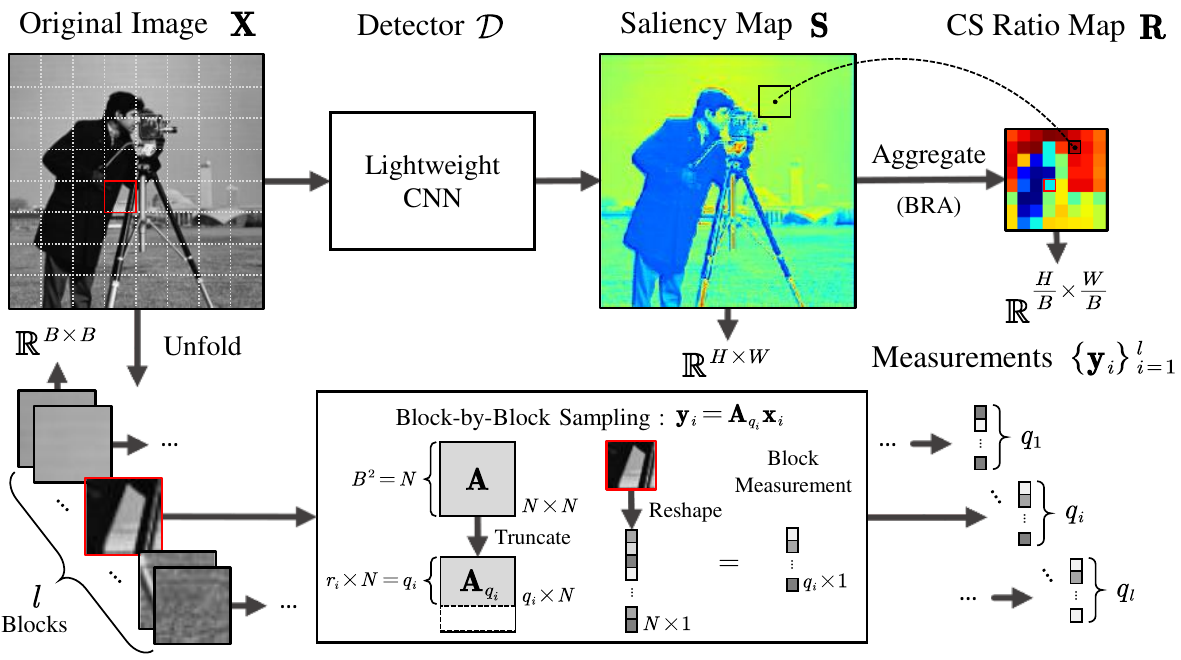}
		\vspace{-20pt}
		\caption{Illustration of the sampling subnet (SS), which conducts an three-stage sampling process consisted of saliency detection, CS ratio allocation, and block-by-block sampling. Each $B\times B$ block is reshaped (or vectorized) to $N\times 1$ and sampled by its corresponding sampling matrix $\mathbf{A}_{q_i}\in \mathbb{R}^{q_i\times N}$.}
		\label{fig: ss}
		\vspace{6pt}
	\end{figure}
	
	\textbf{(3) Deep unfolding reconstruction.} With theoretical guarantees and favorable strong interpretability, deep unfolding technique integrates both optimization-based and network-based methods by fusing the data-fidelity constraints into the learning model. Inspired by the previous works \cite{ISTA-Net, AMP-Net, OPINE-Net, ADMM-Net, you2021coast, chun2020momentum}, our recovery network is built on unfolding framework, maps each optimization iteration into a learnable phase structure, and recovers the target image step-by-step.
	
	\vspace{-6pt}
	\subsection{Architecture Design of CASNet}
	In this subsection, we will illustrate the architecture design of CASNet and related techniques. As Fig. \ref{fig: arch} shows, CASNet is composed of a sampling subnet (SS), an initialization subnet (IS), and a recovery subnet (RS). Note that our CASNet mainly focuses on one-channel natural images, and it could be easily extended\footnote{For colorful image or video data, the CASNet extension could be achieved by sampling and reconstructing channel-by-channel or frame-by-frame.} to colorful image or video CS tasks.
	
	\subsubsection{\textbf{Sampling Subnet (SS)}}
	As Fig. \ref{fig: ss} illustrates, the sampling process in SS can be divided into three stages: saliency detection, CS ratio allocation, and block-by-block sampling.
	
 	\textbf{(1) Saliency detection.} In the first stage, instead of using a manually set detecting method \cite{SaliencyBased, BCS-Net}, we adopt a CNN as the saliency detector $\mathcal{D}$ to evaluate the saliency of each location and highlight the importance information of different regions. It consists of a convolution layer, three residual blocks and another convolution layer to estimate and give out a single-channel saliency map $\mathbf{S}$ with the same size $H\times W$ as input.

        \vspace{-15pt}
	\newcommand\mycommfont[1]{\footnotesize\ttfamily\textcolor{blue}{#1}}
	\SetCommentSty{mycommfont}
	\setlength{\textfloatsep}{0pt}
	\begin{algorithm}
		\label{BRA}
		\caption{Block ratio aggregation (BRA).}
		\LinesNumbered
		\KwIn{Saliency map $\mathbf{S}\in \mathbb{R}^{H\times W}$, block size $B$, target average measurement size $q$ (the expected CS ratio is ${q}/{B^2}$ or $q/N$), upper bound $K$.}
		\KwOut{CS ratio map $\mathbf{R}\in \mathbb{R}^{(H/B)\times (W/B)}$.}
		$\mathbf{S}:=\text{softmax}(\mathbf{S})$\tcp*{softmax normalization}
		$l:=(H/B)\times (W/B)$\tcp*{total number of blocks}
		$\mathbf{Q} := q\times l\times \text{sumpool}_{B\times B}(\mathbf{S})$\;
		$i:=0$, $T:=10$\;
		\While{$\text{true}$}{
			$i:=i+1$\;
			$\mathbf{Q} := \text{round}(\text{clip}_{0,K}(\mathbf{Q}))$\;
			$\delta :=\text{average}(\mathbf{Q}) - q$\;
			\uIf{$\delta$ equals $0$}{
				break\;
			}\uElseIf{$i\le T$}{
				$\mathbf{Q} := \mathbf{Q}-\delta$\tcp*{method \#1}
			}\Else{
				generate a random matrix $\mathbf{\Delta}\in \mathbb{N}^{(H/B)\times (W/B)}$ following the multinomial distribution with parameters $\text{abs}(\delta l)$ and $\mathbf{P}\in \mathbb{R}^{(H/B)\times (W/B)}$, where all elements of $\mathbf{P}$ are set to $(1/l)$\;
				$\mathbf{Q} := \mathbf{Q}-\text{sign}(\delta) \times \mathbf{\Delta}$\tcp*{method \#2}
			}
		}
		$\mathbf{R}:={\mathbf{Q}}/{(B^2)}$\tcp*{final normalization}
		\Return $\mathbf{R}$\tcp*{end of BRA}
	\end{algorithm}
	\vspace{-10pt}
	
	\textbf{(2) CS ratio allocation.} In the second stage, $\mathbf{S}$ is logically divided into $l$ blocks of size $B\times B$, where $B = \sqrt{N}$. Then the block aggregation is performed to get a CS ratio map $\mathbf{R}$ which incorporates the allocated block sampling rates $\left\{r_i\right\} _{i=1}^{l}$. In fact, the available sampling rate of each block can be only selected from $\left\{ {q}/{N} \right\} _{q=1}^{N}$ due to the limited sampling matrix size, where we denote $q_i$ as the measurement size of the $i$-th block. We design a block ratio aggregation (BRA) strategy, which can be summarized as softmax normalization, sumpooling aggregation and error correction, to achieve accurate CS ratio allocation. Concretely, BRA applies softmax normalizer to $\mathbf{S}$, performs $B\times B$ sumpooling to get an aggregated weight map, and times it with the target sum of measurement size $(ql)$ to get the measurement size map $\mathbf{Q}$, then $\mathbf{Q}$ is checked and corrected iteratively. In each correction iteration, $\mathbf{Q}$ is sheared and discretized to make all its elements be integers in $\left[ 0,N \right]$ with specifying the block measurement size upper bound as $N$, then its average error $\delta$ determines whether the correction needs to be performed. Alg. \ref{BRA} exhibits the details of BRA, and numbered lines 11-15 show its two different correction methods: uniform descent and random error elimination based on the multinomial distribution. In our experiments, the BRA strategy is implemented by PyTorch \cite{PyTorch} with the differential property which enables the backpropagations can reach the saliency detector $\mathcal{D}$ and guide the update of its parameters. It distributes $\mathbf{R}$ in no more than 16 correction iterations in all our evaluations. Appx. \ref{instance_BRA} and Appx. \ref{convergence_analysis_BRA} provide a simple instance and our convergence analysis of BRA strategy, respectively.
	
	\textbf{(3) Block-by-block sampling.} In the final stage, the original image $\mathbf{X}$ is unfolded into $B\times B$ blocks $\{\mathbf{x}_i\}_{i=1}^l$ , and each block $\mathbf{x}_i$ is sampled by $\mathbf{y}_i=\mathbf{A}_{q_i}\mathbf{x}_i$ with its corresponding sampling matrix $\mathbf{A}_{q_i}$. Note that $\mathbf{A}_{q_i}$ is obtained by truncating $\mathbf{A}$ and preserving its first $q_i$ rows, here $q_i=r_i\times N$, and $r_i$ is the corresponding allocated CS ratio in $\mathbf{R}$.
	
	\begin{figure}[!h]
         \vspace{-5pt}
		\centering
		\includegraphics[width=0.48\textwidth]{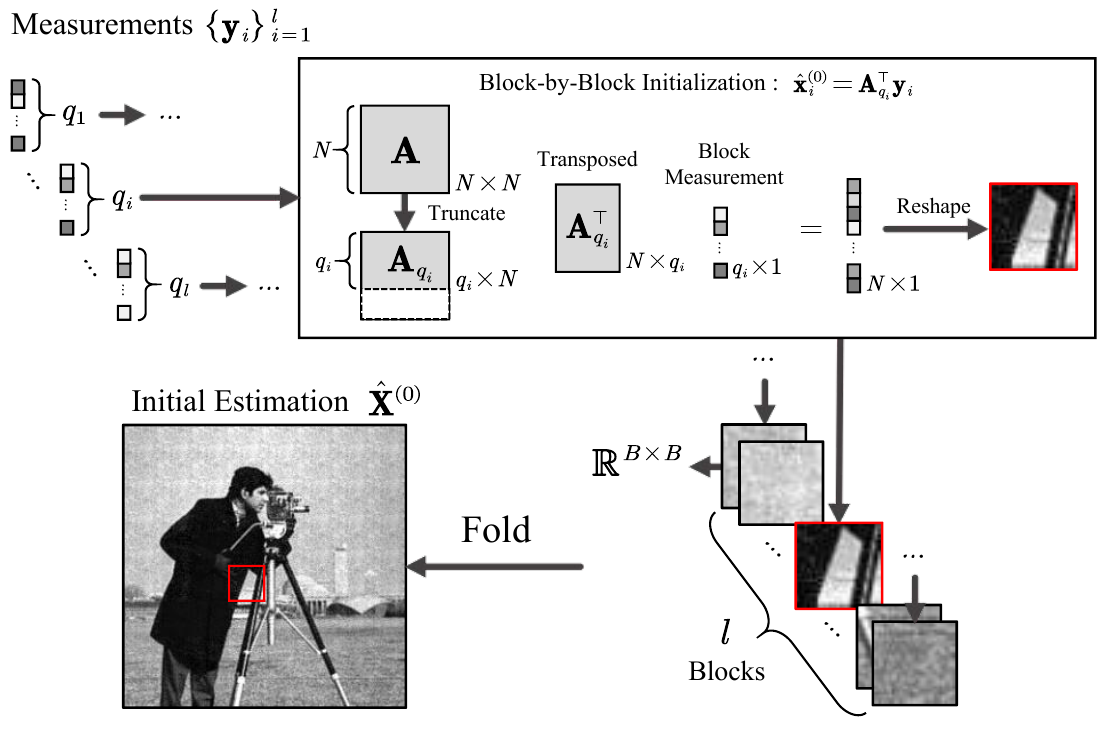}
		\vspace{-12pt}
		\caption{Illustration of the initialization subnet (IS), which reuses the sampling matrices to handle the blocks-measurements dimensionality mismatch.}
		\label{fig: is}
		\vspace{-5pt}
	\end{figure}
	
	\subsubsection{\textbf{Initialization Subnet (IS)}}
	Fig.~\ref{fig: is} illustrates the initialization process in IS. Instead of exploiting a fully-connected layer to handle the dimensionality mismatch between the image blocks and their measurements \cite{DL2_ReconNet, AMP-Net}, following \cite{mousavi2017learning}, the IS directly uses $\hat{\mathbf{x}}^{(0)}_i=\mathbf{A}_{q_i}^\top \mathbf{y}_i$ for block initialization. Concretely, each block is initialized by its corresponding transposed sampling matrix $\mathbf{A}_{q_i}^\top$, then all results are reshaped and folded to form the initial estimation $\mathbf{\hat{X}}^{\left( 0 \right)}$. Without introducing parameters, IS is a simple but fast and efficient implementation to bridge SS and the recovery subnet, and can be easily applied to multi-block processing tasks with different CS ratios.
	
	\begin{figure*}[t]
		\centering
		\includegraphics[width=1.00\textwidth]{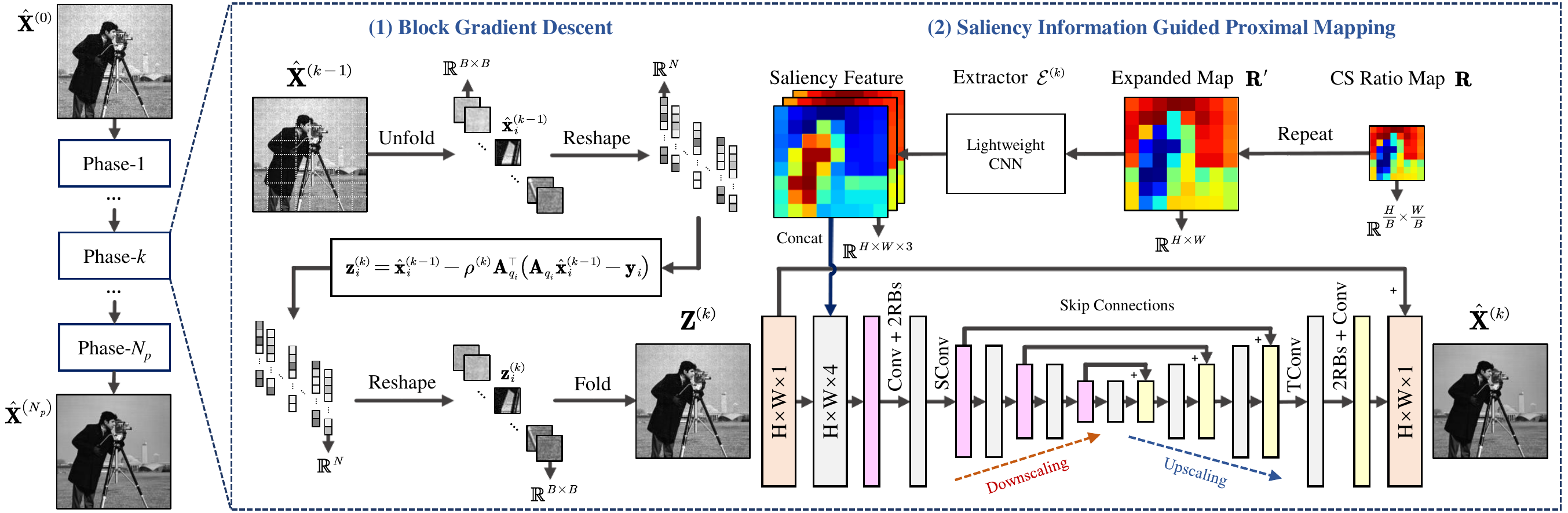}
		\vspace{-20pt}
		\caption{Illustration of the recovery subnet (RS), which conducts a two-stage recovery process consisting of block gradient descent and saliency information guided proximal mapping. The blocks are processed individually in the first stage and then folded to be jointly recovered in the second stage to address the blocking artifacts by exploiting the CS ratio map and perceiving the inter-block relationships. In the proximal mapping part (\textcolor{blue}{right}), ``2RBs" denotes two residual blocks in series, and the pink and yellow bars correspond to the input and output features of the U-Net encoder and decoder blocks, respectively.}
		\label{fig: rs}
		\vspace{-14pt}
	\end{figure*}
	
	\subsubsection{\textbf{Recovery Subnet (RS)}}
	Considering the simplicity and interpretability, we follow \cite{lefkimmiatis2018universal} and directly unfold the traditional proximal gradient descent (PGD) \cite{parikh2014proximal} which solves Eq. (\ref{traditional_optimization_problem}) by iterating between the following two update steps:
	
	\begin{equation}
		\label{z_module}
		\setlength{\abovedisplayskip}{-5pt}
		\setlength{\belowdisplayskip}{-2pt}
		\begin{split}
			\mathbf{z}^{(k)} &= \mathbf{\hat{x}}^{(k-1)} - \rho \mathbf{\Phi}^{\top} (\mathbf{\Phi} \mathbf{\hat{x}}^{(k-1)} - \mathbf{y}) ,
		\end{split}
	\end{equation}
	
	\begin{equation}
		\label{x_module}
		\setlength{\abovedisplayskip}{-3pt}
		\setlength{\belowdisplayskip}{5pt}
		\begin{split}
			\mathbf{\hat{x}}^{\left( k \right)}=\underset{\mathbf{{x}}}{\text{arg}\min}\frac{1}{2}\lVert \mathbf{{x}}-\mathbf{z}^{\left( k \right)} \rVert _{2}^{2}+\lambda \mathcal{R}\left( \mathbf{{x}} \right) ,
		\end{split}
	\end{equation}
	where $k$ denotes the PGD iteration index, and $\rho$ is the step size. Here Eq. (\ref{z_module}) is a trivial gradient descent step, while Eq. (\ref{x_module}) is the so-called proximal mapping. It is worth noting that CASNet is not limited with PGD and may be extended to other optimization algorithms like the iterative shrinkage-thresholding algorithm (ISTA) \cite{ISTA} and half quadratic splitting (HQS) \cite{geman1995nonlinear}. As Fig. \ref{fig: rs} shows, RS is composed of $N_p$ phases, each phase conducts a two-stage process consisted of block gradient descent and saliency information guided proximal mapping, which correspond to the two update PGD steps.
	
	\textbf{(1) Block gradient descent.} In the first stage, we directly map the first step of the PGD iteration and define the block gradient descent according to Eq. (\ref{z_module}). By introducing the learnable step size $\rho ^{(k)}$, this stage can be expressed as:
	
	\begin{equation}
		\label{z_module_RS}
		\setlength{\abovedisplayskip}{-8pt}
		\setlength{\belowdisplayskip}{5pt}
		\begin{split}
			\mathbf{z}^{(k)}_i &= \mathbf{\hat{x}}^{(k-1)}_i - \rho ^{(k)} \mathbf{A}_{q_i}^{\top} (\mathbf{A}_{q_i} \mathbf{\hat{x}}^{(k-1)}_i - \mathbf{y}_i) .
		\end{split}
	\end{equation}
	
	In this stage, the image data $\mathbf{\hat{X}}^{(k-1)}$ is unfolded into $l$ blocks and each block is processed individually. Then the intermediate results of all blocks are folded to form $\mathbf{Z}^{(k)}$ and sent to the next stage for joint recovery with perceiving the inter-block relationships and addressing the blocking artifacts.
	
	\textbf{(2) Saliency information guided proximal mapping.} Following \cite{DPIR} that proposes a powerful denoiser exhibiting flexibility and effectivity in image restoration tasks, we propose to conduct a CNN-based proximal mapping to solve Eq.~(\ref{x_module}) with the exploitation of CS ratio information. As Fig.~\ref{fig: rs} illustrates, the CS ratio map $\mathbf{R}\in\mathbb{R}^{(H/B)\times (W/B)}$ is repeated to obtain an expanded map $\mathbf{R}^{'}\in\mathbb{R}^{H\times W}$ by filling the elements of $\mathbf{R}^{'}$ of different image blocks with their corresponding CS ratios in $\mathbf{R}$. Then a data-driven extractor that consists of a convolution layer, three residual blocks and another convolution layer with $1\times 1$ kernels are used to embed $\mathbf{R}^{'}$ into a three-dimensional feature space. By concatenating  $\mathbf{Z}^{(k)}$ with the saliency feature, we propose a powerful proximal mapping network providing a flexible way of handling different CS ratios by taking the concatenated feature as input and giving out the recovered residual content. Here we denote the feature extractor and the proximal mapping network as $\mathcal{E}^{\left( k \right)}$ and $\mathcal{P}^{\left( k \right)}$ respectively, then this process can be formulated as:
	
	\begin{equation}
		\label{x_module_RS}
		\setlength{\abovedisplayskip}{-8pt}
		\setlength{\belowdisplayskip}{5pt}
		\begin{split}
			\mathbf{\hat{X}}^{\left( k \right)}=\mathbf{Z}^{\left( k \right)}+\mathcal{P}^{\left( k \right)}\left( \left[ \mathbf{Z}^{\left( k \right)}\text{ }|\text{ }\mathcal{E}^{\left( k \right)}\left( \mathbf{R}^{'} \right) \right] \right),
		\end{split}
	\end{equation}
	where $[\mathbf{F}_1|\mathbf{F}_2]$ is the concatenation of two feature maps $\mathbf{F}_1$ and $\mathbf{F}_2$ with the spatial size $H\times W$ in the channel dimension.
	
	Each $\mathcal{P}^{\left( k \right)}$ adopts a U-shaped structure like \cite{UNet, DPIR} with four scales. It consists of three encoder blocks and three decoder blocks. Each encoder block is stacked by a convolution layer, two residual blocks, and a $2\times 2$ strided convolution (SConv) layer, and each decoder block consists of a $2\times 2$ transposed convolution (TConv) layer, two residual blocks, and a convolution layer. There are three skip connections for the last three scales, and the feature channel numbers of the four scales are set to $\{16,32,64,128\}$, respectively.
	
	\vspace{-8pt}
	\subsection{Model Learning and Boosting}
	\label{training_scheme}
	\textbf{Trainable Components:} In light of previous descriptions, our main ideas in Sec.~\ref{main_ideas} can be successfully implemented and mapped into CASNet. Concretely, the learnable parameter set in CASNet, denoted by $\mathbf{\Theta}$, includes the saliency detector $\mathcal{D}$ in SS, generating matrix $\mathbf{A}$, step sizes $\rho ^{(k)}$, saliency feature extractors $\mathcal{E} ^{(k)}$ and proximal mapping networks $\mathcal{P} ^{(k)}$ in RS, \textit{i.e.}, $\mathbf{\Theta} = \left\{ \mathcal{D}, \mathbf{A}\right\} \cup   \left\{ \rho  ^{(k)}, \mathcal{E}  ^{(k)}, \mathcal{P}  ^{(k)} \right\}_{k=1}^{N_p}$.
	
	\textbf{Learning Objective:} Given the training set $\left\{ \mathbf{X}_i \right\} _{i=1}^{N_b}$ with $N_b$ patches of size $B\sqrt{l} \times B\sqrt{l}$, by taking $\mathbf{X}_i$ and a non-negative integer $q_i$ as inputs, we aim to reduce the discrepancy between $\mathbf{X}_i$ and the sampled and recovered $\mathcal{F}_{\text{CASNet}}\left( \mathbf{X}_i,q_i \right)$, where $q_i$ represents the average size of block measurements, corresponds to the target CS ratio ${q_i}/{N}$ and is randomly selected from $\left\{ 1,2,\cdots ,N \right\}$ for each iteration. Due to the well-defined structure and the tight coupling among different components, we employ the following $\ell_1$-loss to train CASNet:
	
	\begin{equation}
		\label{loss}
		\setlength{\abovedisplayskip}{-9pt}
		\setlength{\belowdisplayskip}{3pt}
		\begin{split}
			\mathcal{L}\left( \mathbf{\Theta} \right) =\frac{1}{lNN_b}\sum_{i=1}^{N_b}{\lVert \mathcal{F}_{\text{CASNet}}\left( \mathbf{X}_i,q_i ;\mathbf{\Theta}\right) -\mathbf{X}_i \rVert _1} .
		\end{split}
	\end{equation}
	
	\noindent In addition, we find that the $\ell_2$-loss replacing $\lVert\cdot\lVert_1$ in Eq. (\ref{loss}) with $\lVert\cdot\lVert_2^2$ also leads to stable convergence and similar recovery accuracies (please refer to our comparison in Tab. \ref{tab:ablation} (12)-(13)).
	
	\textbf{Initialization Scheme for the Generating Matrix $\mathbf{A}$:} To accelerate the training convergence, instead of using random initialization methods, we propose to initialize the generating matrix based on the singular value decomposition (SVD). Specifically, by partitioning each training patch $\mathbf{X}_i$ into $l$ blocks to obtain $\left\{ \mathbf{x}_i \right\} _{i=1}^{lN_b}$, we aim to get a generating matrix initialization $\mathbf{A}_{\text{init}}$ which satisfies the condition that for any sampling matrix $\mathbf{A}_{q}$, there is \textcolor{blue}{$\mathbf{A}_{q}=\arg\min_{\mathbf{A}}\lVert \mathbf{A}^{\top}\mathbf{AD}-\mathbf{D} \rVert _{F}^{2}$}, where $\mathbf{D}=\left[ \mathbf{x}_1,\cdots, \mathbf{x}_{lN_b} \right]$ has an SVD of $\mathbf{D}=\mathbf{U\Sigma V}^{\top}$ with singular values $\sigma _{i}\equiv \mathbf{\Sigma }_{i,i}$ satisfying $\sigma _1\ge \sigma _2\ge \cdots \ge \sigma _{N}$. The above cost function marked in \textcolor{blue}{blue} is to minimize the distance between the original training image blocks and their corresponding sampled and initialized ones by any $\{\mathbf{A}_q,\mathbf{A}_q^\top\}$ from $\mathbf{A}_{\text{init}}$. According to the Eckart–Young–Mirsky theorem \cite{eckart1936approximation}, the optimal $\mathbf{A}_{\text{init}}$ is equal to $\mathbf{U}^\top$. So before training, we perform the above SVD and initialize $\mathbf{A}$ with $\mathbf{A}_{\text{init}}=\mathbf{U}^\top$.
	
	\textbf{Random Transformation Enhancement (RTE) Strategy:} To improve the training efficiency and network robustness, we propose an RTE strategy by introducing randomness into RS. Before obtaining saliency feature, the $k$-th phase applies a geometric transformation $\mathcal{H}^{(k)}$ to $\mathbf{R}^{'}$ and $\mathbf{Z}^{(k)}$, and performs the corresponding inverse $ \widetilde{\mathcal{H}}^{(k)}$ on the phase output. With RTE, the recovery process of Eq. (\ref{x_module_RS}) can be reformulated as follows:
	
	\begin{equation}
		\label{x_module_RS_reformulated}
		\setlength{\abovedisplayskip}{-4pt}
		\setlength{\belowdisplayskip}{5pt}
		\begin{split}
			\mathbf{J}^{\left( k \right)}&=\mathcal{H}^{\left( k \right)}\left( \mathbf{Z}^{\left( k \right)} \right) , \mathbf{W}^{\left( k \right)}= \mathcal{E}^{\left( k \right)}\left( \mathcal{H}^{\left( k \right)}\left( \mathbf{R}^{'} \right) \right) , \\
			\mathbf{\hat{X}}^{\left( k \right)}&=\widetilde{\mathcal{H}}^{\left( k \right)}\left( \mathbf{J}^{\left( k \right)}+\mathcal{P}^{\left( k \right)}\left( \left[ \mathbf{J}^{\left( k \right)} \text{ }|\text{ }\mathbf{W}^{\left( k \right)} \right] \right) \right) ,
		\end{split}
	\end{equation}
	where $\mathcal{H}^{\left( k \right)}$ is randomly choosed from eight transforms including rotations, flippings and their combinations \cite{ADNet}. RTE is to make better use of bottlenecks between each two adjacent phases by enhancing rotation/flipping invariance of phase outputs and enforcing them to have similar properties to images.
	
	\begin{figure*}[t]
		\centering
		\hspace{-6.5pt}
		\vspace{-6pt}
		\includegraphics[width=1.005\textwidth]{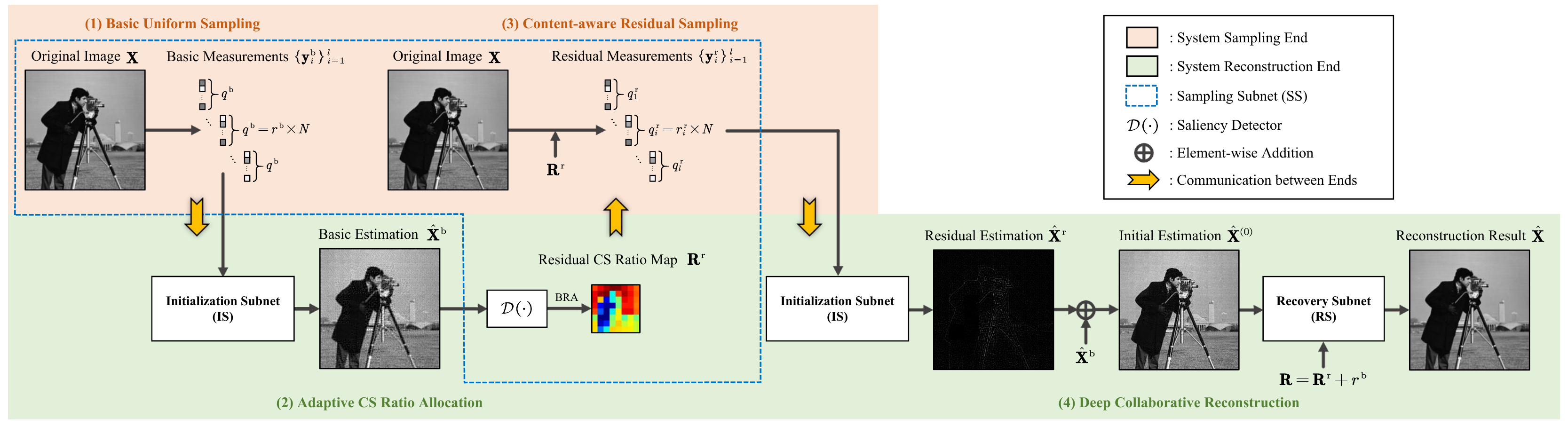}
		\vspace{-18pt}
		\caption{Illustration of our proposed CASNet implementation, which is designed to be an interface between CASNet and the physical CS system and driven by a four-stage pipeline consisted of basic uniform sampling, adaptive CS ratio allocation, content-aware residual sampling and deep collaborative reconstruction.}
		\label{fig: imple}
		\vspace{-12pt}
	\end{figure*}
	
	\vspace{-8pt}
	\subsection{Model Evaluation and Deployment}
	As a comprehensive and general CS framework, CASNet is designed to support various deployment schemes in practical scenarios with different requirements. But there may be still a distance between CASNet itself and real-world applications since it contains the ideal saliency-based sampling process by default, which uses the clean image to produce a saliency map. However, it is not always possible to directly access the complete signal information before CS sampling in some practical deployments. For example, in the context of single-pixel imaging \cite{SinglePixelImaging1, SinglePixelImaging2}, we could not scan the original image to obtain a CS ratio map $\mathbf{R}$ since the image information is unknown (\textit{i.e.} it is not sampled). In this subsection, we will further provide a common CASNet implementation based on a simple system model which physically consists of a \textit{sampling end} and a \textit{reconstruction end}. As Fig. \ref{fig: imple} illustrates, its processing pipeline can be divided into four parts: basic uniform sampling, adaptive CS ratio allocation, content-aware residual sampling and deep collaborative reconstruction.
	
	\textbf{(1) Basic uniform sampling.} Considering that the CASNet may not directly access the complete signal information for saliency detection and CS ratio allocation before its sampling in practical evaluations. Here we propose to first perform a trivial basic uniform sampling which uses the first few rows of the learned $\mathbf{A}$ to obtain some measurements to drive the following saliency-based allocation process and ensure the content-aware property. In the first stage, with presetting the expected CS ratio $r$ and a basic sampling proportion denoted as $\gamma\in[0,1]$, the system sampling end uniformly samples the original image signal with the basic CS ratio $r^\text{b}=\gamma \times r$ by $\mathbf{A}_{q^\text{b}} (q^\text{b}=r^\text{b}\times N)$ generated from $\mathbf{A}$ and sends the basic measurements $\left\{ \mathbf{y}_i^\text{b} | \mathbf{y}_i^\text{b}= \mathbf{A}_{q^\text{b}} \mathbf{x}_i \right\}_{i=1}^{l}$ to the reconstruction end.
	
	\textbf{(2) Adaptive CS ratio allocation.} In the second stage, the system reconstruction end adopts the fast and efficient IS to bring the received measurements back to the image domain by $\mathbf{\hat{x}}_i^\text{b}=\mathbf{A}_{q^\text{b}}^\top \mathbf{y}_i^\text{b}$, then folds them to form the basic estimation $\mathbf{\hat{X}}^\text{b}\in\mathbb{R}^{H\times W}$ and perform the residual CS ratio allocation based on the saliency map $\mathbf{S}=\mathcal{D}(\mathbf{\hat{X}}^\text{b})$ and BRA with the target average measurement size of $(r\times N-q^\text{b})$ and the corrected upper bound $(N-q^\text{b})$ of block measurement size to obtain the residual CS ratio map $\mathbf{R}^\text{r}$ and send it to the sampling end as a key part of the secondary sampling request.
	
	\textbf{(3) Content-aware residual sampling.} In the third stage, the system sampling end achieves the network content-aware property by the non-uniform residual sampling based on the received $\mathbf{R}^\text{r}$. Each block $\mathbf{x}_i$ is sampled by the residual sampling matrix slice $ \mathbf{A}_{q^\text{b}+1, q_i}=\mathbf{A}\left[ q^\text{b}+1:q_i\right] (q_i=q^\text{b}+r_i^\text{r}\times N)$, and the residual measurements $\left\{ \mathbf{y}_i^\text{r} | \mathbf{y}_i^\text{r}=\mathbf{A}_{q^\text{b}+1, q_i} \mathbf{x}_i \right\}_{i=1}^{l}$ are then sent to the reconstruction end.
	
	\textbf{(4) Deep collaborative reconstruction.} In the last stage, the reconstruction end employs IS again to obtain the block residual initialization $\mathbf{\hat{x}}_i^\text{r}=\mathbf{A}_{q^\text{b}+1, q_i}^\top \mathbf{y}_i^\text{r}$, folds them to form the residual estimation $\mathbf{\hat{X}}^\text{r}$ and gives out the initial estimation by $\mathbf{\hat{X}}^{\left( 0 \right)} = \mathbf{\hat{X}}^\text{b}+\mathbf{\hat{X}}^\text{r}$. Note that each block $\mathbf{x}_i$ is equivalent to be sampled and initialized by its corresponding complete sampling matrix $\mathbf{A}_{q_i}$. Then $\mathbf{\hat{X}}^{\left( 0 \right)}$ and the corrected CS ratio map $\mathbf{R}$ obtained by adding $r^\text{b}$ to each entry of $\mathbf{R}^\text{r}$ are sent into RS to get the collaborative reconstruction $\mathbf{\hat{X}}=\mathbf{\hat{X}}^{(N_p)}$.
	
	As we can see, the above four-stage implementation focuses on the sampling-initialization process and eliminates the gap between CASNet and the physical CS system by acquiring a proportion of measurements in a uniform sampling manner to achieve the saliency-based CS ratio allocation and content-aware residual sampling. Under this scheme, SS is distributed at both the system sampling end and reconstruction end (see Fig.~\ref{fig: imple}), while IS and RS are deployed at the reconstruction end. An appropriate value of the basic sampling proportion $\gamma$ needs to be determined in a specific real-world scenario as it directly controls the trade-off between the signal pre-knowledge sufficiency for saliency evaluation and the CS ratio allocating space. Furthermore, we note that our CASNet is not limited to the proposed scheme, it may also support many other implementations such as the schemes mentioned in \cite{SaliencyBased, BCS-Net} which adopt a low-resolution complementary sensor to acquire a sample image to generate a saliency map of the scene under view. For keeping the training simplicity and fair evaluations in our experiments, we train CASNet based on the simple pipeline connected by the three subnets in series with a unidirectional information flow (see Fig.~\ref{fig: arch}) and the learning and boosting strategies in Sec.~\ref{training_scheme} with using the complete image for saliency detections and samplings, and only test it under our system implementation described in this subsection.
	
	\vspace{-6pt}
	\section{Experimental Results}
	\vspace{-2pt}
	\subsection{Implementation Details}
	Following \cite{DL3_CSNet, SCSNet, BCS-Net}, we set the block size $B=32$ and $N=1024$, choose the basic sampling proportion $\gamma=0.2822$, and extract luminance components of 25600 randomly cropped $128\times 128$ image patches from T91 \cite{dong2014learning} and Train400 \cite{DnCNN}, \textit{i.e.}, $N_b=25600$ and $l=16$. All residual blocks adopt the classic Conv-ReLU-Conv structure with an identity connection \cite{ResNet}. The convolution layers in $\mathcal{D}$ and $\mathcal{P}^{(k)}$ use $3\times 3$ kernels, and the intermediate feature channel numbers of $\mathcal{D}$ and $\mathcal{E}^{(k)}$ are set to 32 and 8, respectively. Except for $\mathcal{E}^{(k)}$, all learnable components are empirically set to be bias-free, \textit{i.e.}, there are no bias used in $\mathbf{A}$ and all $3\times 3$ convolution layers.
	
	We implement CASNet with PyTorch \cite{PyTorch} on a Tesla V100 GPU, employ Adam \cite{Adam} optimizer with a momentum of 0.9 and a weight decay of 0.999, and adopt a batch size of 64. It takes about five days to train a 13-phase CASNet for 300 epochs with a learning rate of $ 1\times10^{-4} $ and 20 fine-tuning epochs with a learning rate of $ 1\times10^{-5} $. Two widely used benchmarks: Set11 \cite{DL2_ReconNet} and CBSD68 \cite{CBSD68} are utilized for test, and all the recovered results are evaluated with the peak signal-to-noise ratio (PSNR) and structural similarity index measure (SSIM) \cite{SSIM} on the Y channel.
	
	\begin{table}
		\centering
		\caption{The experiments of ablation studies conducted with five CS ratios on Set11 \cite{DL2_ReconNet}. The best results are labeled in \textbf{bold}.}
		\hspace{-4pt}
		\vspace{-10pt}
		\resizebox{0.49\textwidth}{!}{%
			\label{tab:ablation}
			\begin{tabular}{|cl|p{0.5cm}p{0.5cm}p{0.5cm}p{0.5cm}p{0.6cm}|l|}
				\hline
				\multicolumn{2}{|c|}{\multirow{2}{*}{Setting}}& \multicolumn{5}{c|}{CS Ratio}& \multirow{2}*{\#Param.}\\ \cline{3-7}
				&&1\%&4\%&10\%&25\%&50\%&\\
				\hline
				\hline
				(1)&Uniform Sampling&21.68&26.14&30.10&35.42&40.69&16.839M\\
				(2)&w/o Saliency Information&21.74&26.25&30.23&35.49&40.79&16.883M\\
				\hline
				\hline
				(3)&Random $\mathbf{A}$ Initialization &21.85&26.36&30.33&35.56&40.84&16.895M\\
				(4)&w/o RTE&21.81&26.25&30.15&35.44&40.79&16.895M\\
				\hline
				\hline
				(5)&$\rho^{(k)}$-Shared&21.93&26.40&30.34&35.63&40.93&16.895M\\
				(6)&$\mathcal{E}^{(k)}$-Shared&21.90&26.35&30.31&35.58&40.89&16.889M\\
				(7)&$\mathcal{P}^{(k)}$-Shared&21.85&26.24&30.29&35.50&40.85&2.325M\\
				(8)&Phase-Shared&21.85&26.22&30.29&35.48&40.84&\textbf{2.319M}\\
				\hline
				\hline
				(9)&Replace $\mathcal{E}^{(k)}$ with $\mathcal{E}_1^{(k)}$&21.83&26.28&30.27&35.56&40.87&16.885M\\
				(10)&Replace $\mathcal{E}^{(k)}$ with $\mathcal{E}_{8}^{(k)}$&\textbf{21.97}&26.37&30.35&35.67&40.91&16.985M\\
				(11)&Replace $\mathcal{E}^{(k)}$ with $\mathcal{E}_{\text{bias-free}}^{(k)}$&21.80&26.15&30.25&35.36&40.56&16.894M\\
				\hline
				\hline
				(12)&Trained with $\ell_2$-loss&21.92&\textbf{26.41}&\textbf{30.52}&\textbf{35.70}&\textbf{41.01}&16.895M\\
				\hline
				\hline
				(13)&Ours&\textbf{21.97}&\textbf{26.41}&30.36&35.67&40.93&16.895M\\
				\hline
		\end{tabular}}
	\end{table}
	
	\begin{table}
		\centering
		\caption{Detailed structural configurations of four different data-driven saliency feature extractors for each phase.}
		\hspace{-4pt}
		\vspace{-10pt}
		\resizebox{0.49\textwidth}{!}{%
			\label{tab:e_structure}
			\begin{tabular}{|c|c|c|c|c|}
				\hline
				Setting&Extractor&Structure&Bias-free&\#Param.\\
				\hline
				\hline
				(9)&$\mathcal{E}_1^{(k)}$&$\text{Conv}^{1\times 1}_{1, 1}$&$\times$&\textbf{2}\\		
				(10)&$\mathcal{E}_8^{(k)}$ &$\text{Conv}^{1\times 1}_{1, 32}+3\times \mathcal{T}^{1\times 1}_{32}+\text{Conv}^{1\times 1}_{32, 8}$ &$\times$&6664\\
				(11)&$\mathcal{E}_\text{bias-free}^{(k)}$ &$\text{Conv}^{1\times 1}_{1, 8}+3\times \mathcal{T}^{1\times 1}_{8}+\text{Conv}^{1\times 1}_{8, 3}$ &\checkmark&416\\
				\hline
				\hline
				(13)&$\mathcal{E}^{(k)}$ (Ours)&$\text{Conv}^{1\times 1}_{1, 8}+3\times \mathcal{T}^{1\times 1}_{8}+\text{Conv}^{1\times 1}_{8, 3}$ &$\times$&475\\
				\hline
		\end{tabular}}
		\vspace{18pt}
	\end{table}
	
	\vspace{-8pt}
	\subsection{Ablation Studies and Discussions}
\begin{figure}
		\hspace{-2pt}
		\begin{minipage}{0.33\textwidth}
		\includegraphics[width=1.0\textwidth]{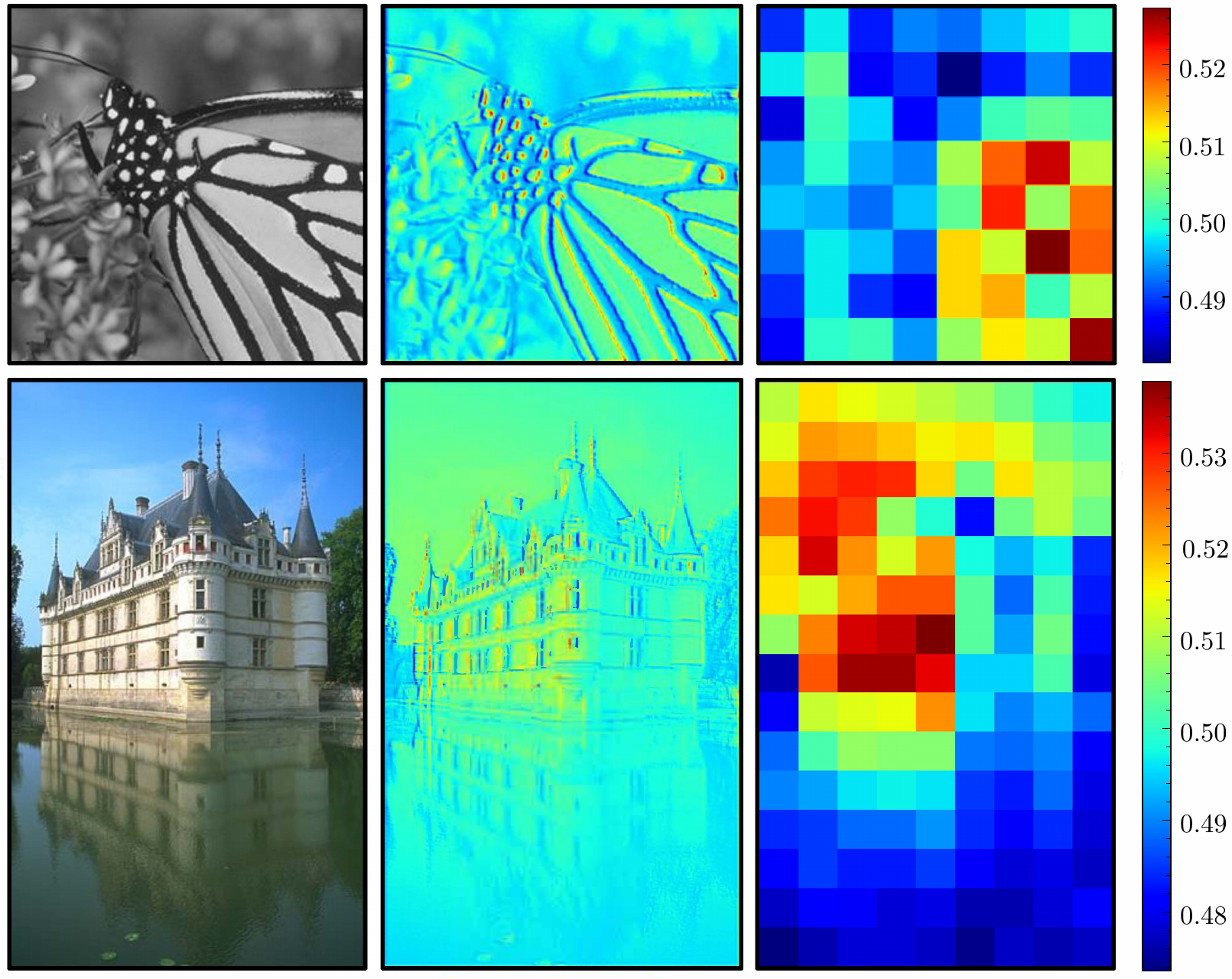}
		\end{minipage}
		\begin{minipage}{0.15\textwidth}
		\includegraphics[width=1.0\textwidth]{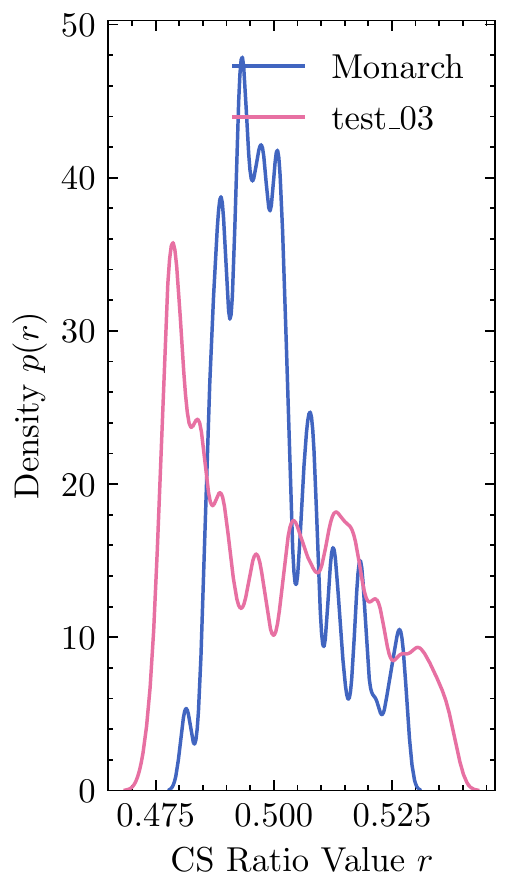}
		\end{minipage}
		\vspace{-10pt}
		\caption{Two saliency-based CS ratio allocation instances on ``Monarch" from Set11 \cite{DL2_ReconNet} (\textcolor{blue}{top left}) and ``test\_03" from CBSD68 \cite{CBSD68} (\textcolor{blue}{bottom left}) with $r=50\%$. Images (\textcolor{blue}{left}) are scanned by $\mathcal{D}$ to obtain saliency maps (\textcolor{blue}{middle left}) and CS ratio maps $\mathbf{R}$ (\textcolor{blue}{middle right}). The corresponding CS ratio distribution curves (\textcolor{blue}{right}) exhibit the instance-wise adaptibility of our BRA strategy.}
		\label{fig: alloc}
		\vspace{-10pt}
	\end{figure}

	\textbf{(1) Effect of Block-wise CS Ratio Allocation and Saliency Information Guidance:} As one of the CASNet main ideas, the saliency-based allocating scheme in SS can perform adaptive CS ratio allocations for different blocks. Fig. \ref{fig: alloc} shows that $\mathcal{D}$ can learn to identify the locations with rich details, and the proposed BRA strategy guarantees the allocation precision by cooperating with $\mathbf{A}$. Tab. \ref{tab:ablation} (1) corresponds to the CASNet variant trained and evaluated under a uniform sampling scheme without content-aware property, and exhibits our more efficient allocations with an average PSNR gain of about 0.26dB. As an efficient approach with low cost to exploit CS ratio information by making $\mathcal{P}^{(k)}$ can perceive the sampling rate distribution, the introduction of $\mathbf{R}^{\prime}$ and $\mathcal{E}^{(k)}$ results in an average PSNR gain of 0.18dB by comparing with giving only $\mathbf{Z}^{(k)}$ to $\mathcal{P}^{(k)}$, as we can see in Tab. \ref{tab:ablation} (2).
	
\begin{figure}
		\hspace{-4pt}
		\includegraphics[width=0.49\textwidth]{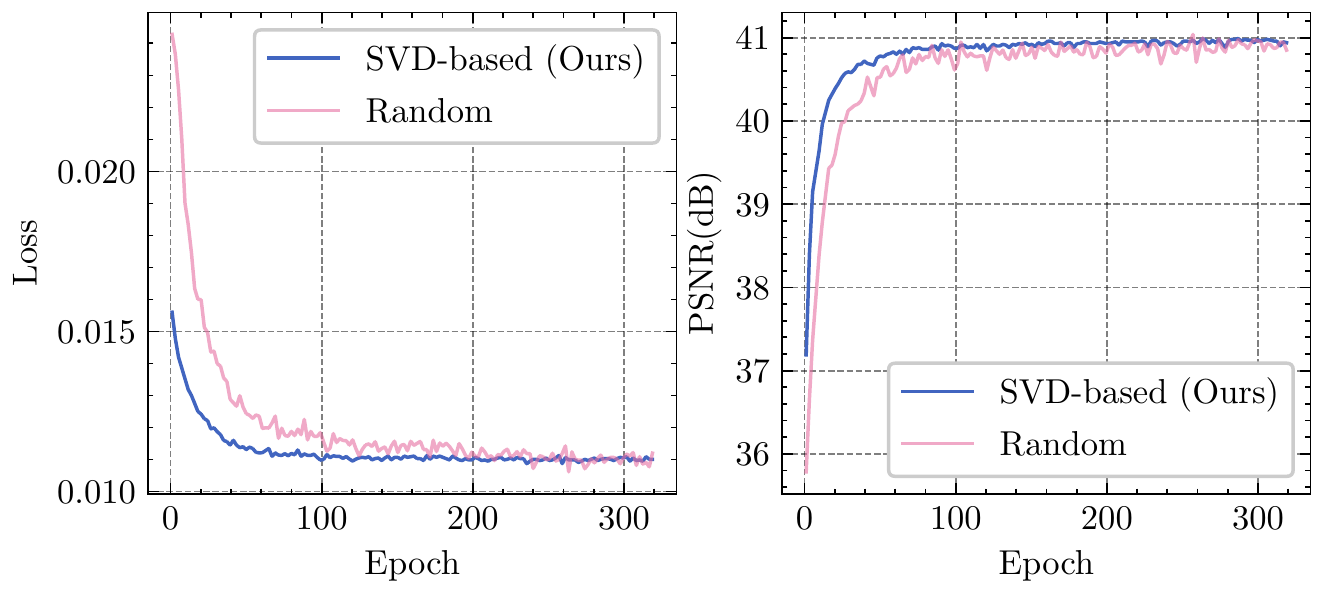}
		\vspace{-22pt}
		\caption{Loss and average PSNR curves with CS ratio $r=50\%$ on Set11 \cite{DL2_ReconNet} achieved by CASNet variants with different initialization methods.}
		\label{fig: train_process}
		\vspace{6pt}
	\end{figure}

	\textbf{(2) Effect of SVD-based Initialization and RTE:} Fig. \ref{fig: train_process} demonstrates the training processes of CASNets with different initialization schemes. Compared with the random initialization \cite{OPINE-Net}, our SVD-based scheme gives a better $\mathbf{A}$ starting point and leads to faster and more stable convergence. As a simple but generalizable enhancement scheme, the RTE strategy is to improve the network robustness by making full use of training data and the inter-phase bottlenecks. These boosting methods are parameter-free but bring 0.08dB and 0.18dB average PSNR gains as Tab. \ref{tab:ablation} (3)-(4) exhibit.

	\textbf{(3) Study of Parameter-sharing Strategies:} We train four CASNet variants with different parameter-sharing strategies as reported by Tab. \ref{tab:ablation} (5)-(8). The most compressed and inflexible variant sharing $\rho ^{(k)}$, $ \mathcal{E} ^{(k)}$ and $ \mathcal{P} ^{(k)}$ among all phases brings the largest average PSNR loss of about 0.13dB. This demonstrates the structural effectiveness and further potential of CASNet in greatly reducing its memory complexity (with about $86.3\%$ parameter number reduction) and being deployed on some lightweight devices with acceptable accuracy drops.
	
\begin{figure}
		\hspace{-4pt}
		\includegraphics[width=0.49\textwidth]{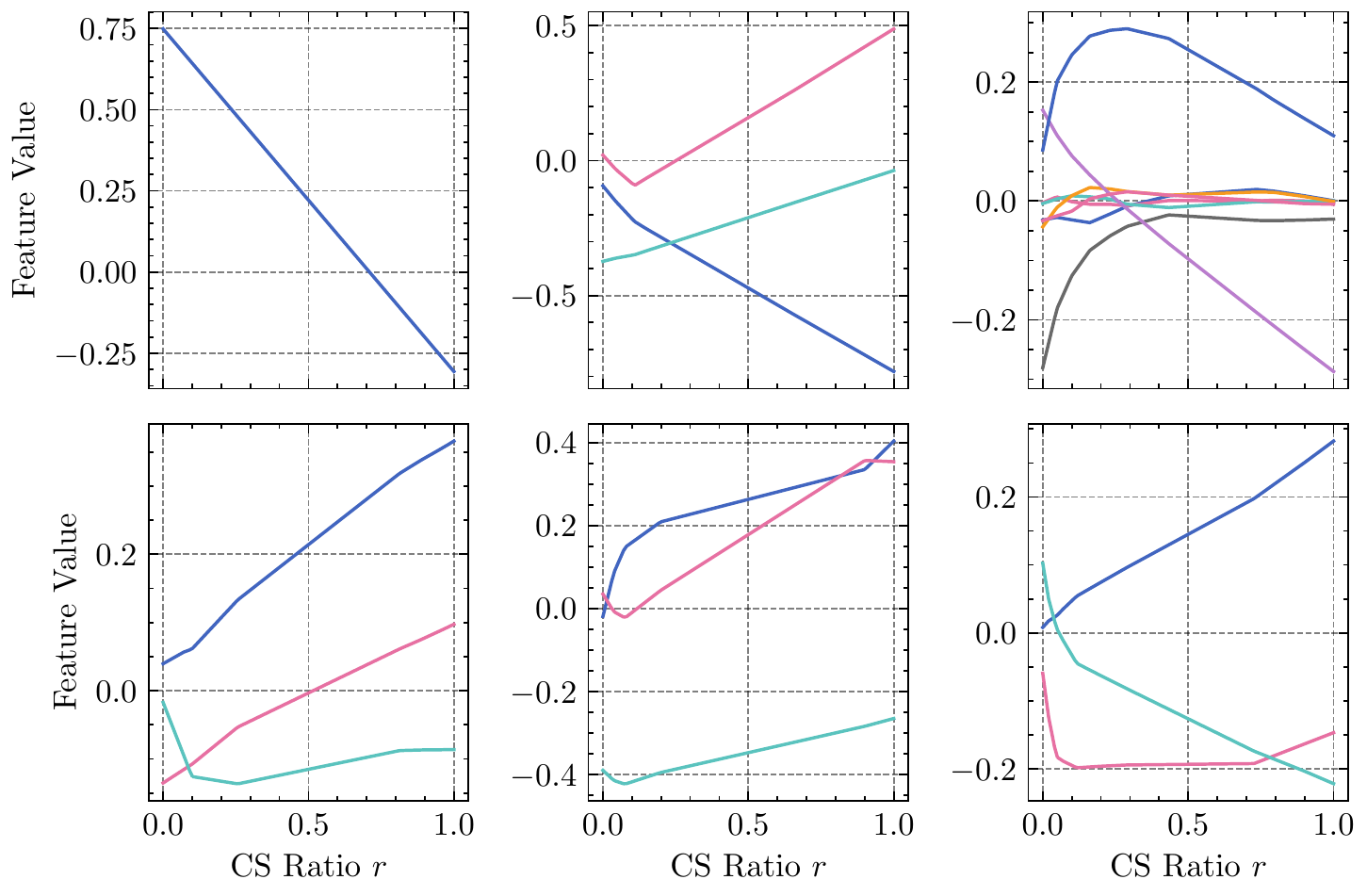}
		\vspace{-20pt}
		\caption{Mappings from CS ratio to saliency feature space done by $\mathcal{E}_1^{(1)}$ (\textcolor{blue}{top left}), $\mathcal{E}^{(1)}$ (\textcolor{blue}{top middle}), $\mathcal{E}_8^{(1)}$ (\textcolor{blue}{top right}), $\mathcal{E}^{(5)}$ (\textcolor{blue}{bottom left}), $\mathcal{E}^{(9)}$ (\textcolor{blue}{bottom middle}) and $\mathcal{E}^{(13)}$ (\textcolor{blue}{bottom right}) in different CASNet variants. The curves in each subgraph correpond to different output feature channels, and the order of which is not concerned in our experiments.}
		\label{fig: e}
		\vspace{-10pt}
	\end{figure}

	\textbf{(4) Study of Saliency Feature Extractor Structure:} As mentioned above, the saliency feature brings the information to guide the proximal mapping network in each phase to recover blocks with non-uniform sampling rates. Instead of manually setting a fixed sampling rate embedding operator, we adopt a data-driven approach and conduct the experiments reported in Tab. \ref{tab:ablation} (9)-(11). Here we denote the convolution layer with $n_o$ kernels of size $n_i\times1\times1$ as $\text{Conv}^{1\times 1}_{n_i, n_o}$, and the residual block with a classic structure of $\text{Conv}^{1\times 1}_{n_f, n_f}$+ReLU+$\text{Conv}^{1\times 1}_{n_f, n_f}$ and an identity skip connection as $\mathcal{T}^{1\times 1}_{n_f}$. The results in Tab. \ref{tab:ablation} corresponding to the extractor structural details provided in Tab. \ref{tab:e_structure} lead to our default setting which only uses a lightweight CNN with 475 parameters including kernel biases to embed the CS ratio into a three-dimensional feature space. Three subgraphs in the first row of Fig.~\ref{fig: e} show the mappings from CS ratio to saliency feature channel values done by (9), (13) and (10) in the first CASNet phases. One can observe that the larger extractor structure of $\mathcal{E}_8^{(k)}$ even brings a weaker performance compared with ours, and we notice that the values in five channels of the output feature given by $\mathcal{E}_8^{(1)}$ have little change as the input varies, this indicates that there are only three channels take the prominent feature information of CS ratio. Three subgraphs in the second row of Fig. \ref{fig: e} illustrate the diversity of mappings done by our default extractors in the 5-th, 9-th, and 13-th phases. These results validate the effectiveness of the proposed extractor $\mathcal{E}^{(k)}$ design and the flexibility of multi-phase framework.

	\begin{figure}
		\hspace{-6pt}
		\includegraphics[width=0.48\textwidth]{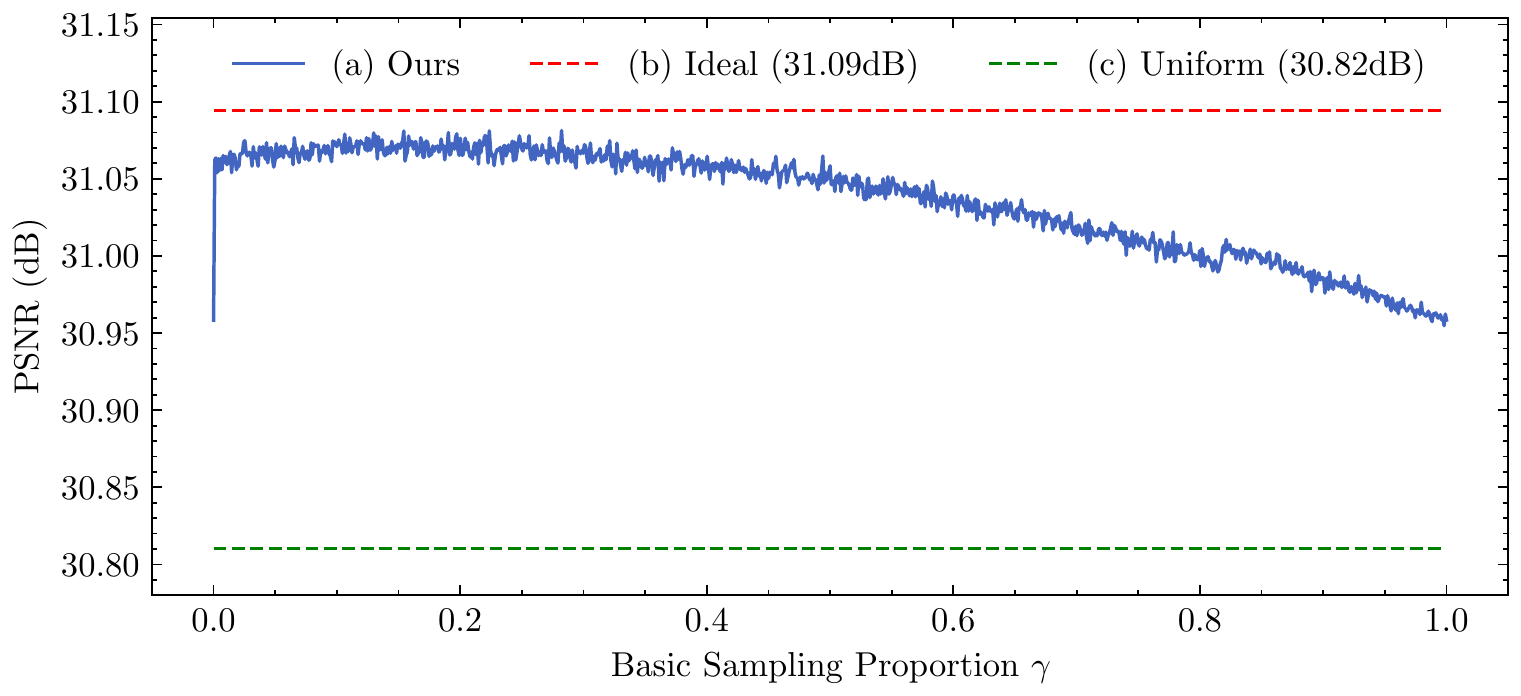}
		\vspace{-8pt}
		\caption{Average PSNR curves on Set11 \cite{DL2_ReconNet} and CBSD68 \cite{CBSD68} with CS ratio $r\in\{1\%, 4\%, 10\%, 25\%, 30\%, 40\%, 50\%\}$ achieved by \textbf{\textcolor{purple}{(a)}} the default CASNet version, \textbf{\textcolor{purple}{(b)}} the ideal version (uses complete image for CS ratio allocation in tests) and \textbf{\textcolor{purple}{(c)}} the uniform sampling version (corresponds to Tab. \ref{tab:ablation} (1)). Our default version \textbf{\textcolor{purple}{(a)}} peaks highest accuracy 31.08dB at $\gamma =0.2822$.}
		\label{fig: gamma}
		\vspace{4pt}
	\end{figure}
	
	\textbf{(5) Study of Basic Sampling Proportion $\gamma$:} As a trade-off controller between the image information completeness and the sampling rate allocating space, the basic sampling proportion $\gamma$ is expected to make CASNet degenerate into a uniform sampling version in evaluation when its value tends to be 0 (lack of signal pre-knowledge) or 1 (lack of residual CS ratio allocating space). Fig.~\ref{fig: gamma} exhibits the average PSNR performances on Set11 \cite{DL2_ReconNet} and CBSD68 \cite{CBSD68} under seven different CS ratios achieved by \textbf{\textcolor{purple}{(a)}} our default CASNet version (trained in content-aware sampling manner), \textbf{\textcolor{purple}{(b)}} the ideal version which employs the complete image for saliency detection and sampling in evaluations (trained in content-aware sampling manner), and \textbf{\textcolor{purple}{(c)}} the trivial uniform sampling version (trained in uniform sampling manner). The default version \textbf{\textcolor{purple}{(a)}} meets its worst performance of 30.95dB when $\gamma$ equals to 0 or 1. And there is a sharp increase (about 0.12dB) at the beginning end of its corresponding curve, which means that only a small proportion of the most important measurements can lead to relatively accurate saliency predictions and allocations. This curve then tends to slowly increase in the range of $[0, 0.2822]$, peaks to 31.08dB at $\gamma =0.2822$, and tends to decreases in $[0.2822, 1]$. We observe that our CASNet implementation \textbf{\textcolor{purple}{(a)}} can even bring a recovery accuracy close to the ideal version \textbf{\textcolor{purple}{(b)}} with a PSNR gap of only about 0.01dB and remain a distance of about 0.26dB compared with the uniform sampling version \textbf{\textcolor{purple}{(c)}}, which is also exceeded by the default degenerated uniform sampling one \textbf{\textcolor{purple}{(a)}} with a PSNR gap of 0.13dB. These results fully verify the effectiveness of our four-stage CASNet implementation and show the importance of keeping the CS ratio diversity in the training process.
	
	\begin{figure}
		\hspace{-6pt}
		\includegraphics[width=0.48\textwidth]{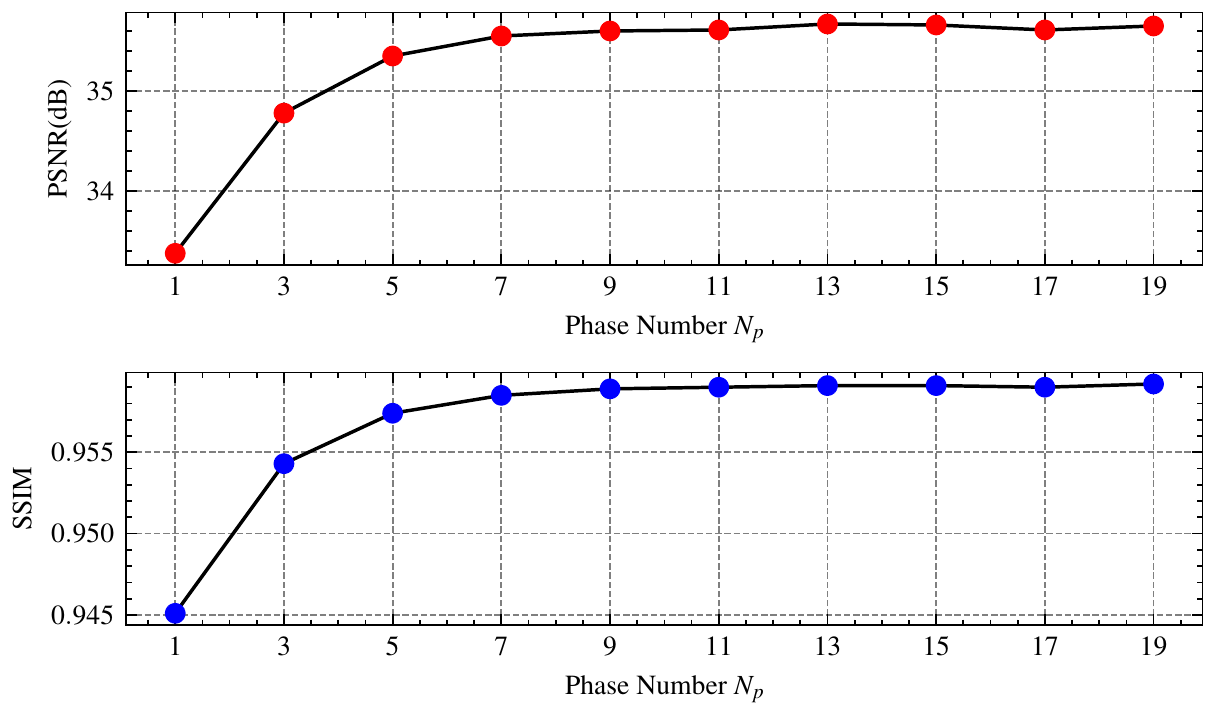}
		\vspace{-8pt}
		\caption{Average PSNR/SSIM curves on Set11 \cite{DL2_ReconNet} achieved by CASNet variants with various phase numbers in the case of CS ratio $r=25\%$.}
		\label{fig: phase_num}
		\vspace{-10pt}
	\end{figure}
	
	\begin{figure}
		\includegraphics[width=0.48\textwidth]{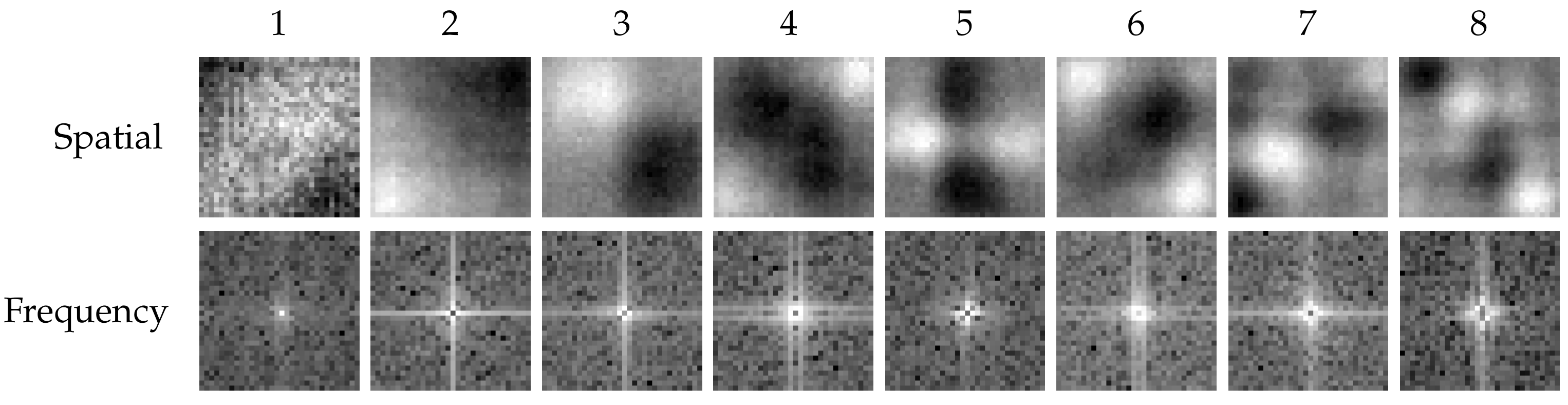}
		\vspace{-6pt}
		\caption{Visualization of the first eight rows of the learned $\mathbf{A}$. The former ones exhibit narrower frequency distributions, which indicates that they pay more attention to low-frequency information.}
		\label{fig: v_phi}
		\vspace{4pt}
	\end{figure}

	\begin{figure*}[t]
		\centering
		\hspace{-8pt}
		\includegraphics[width=1.01\textwidth]{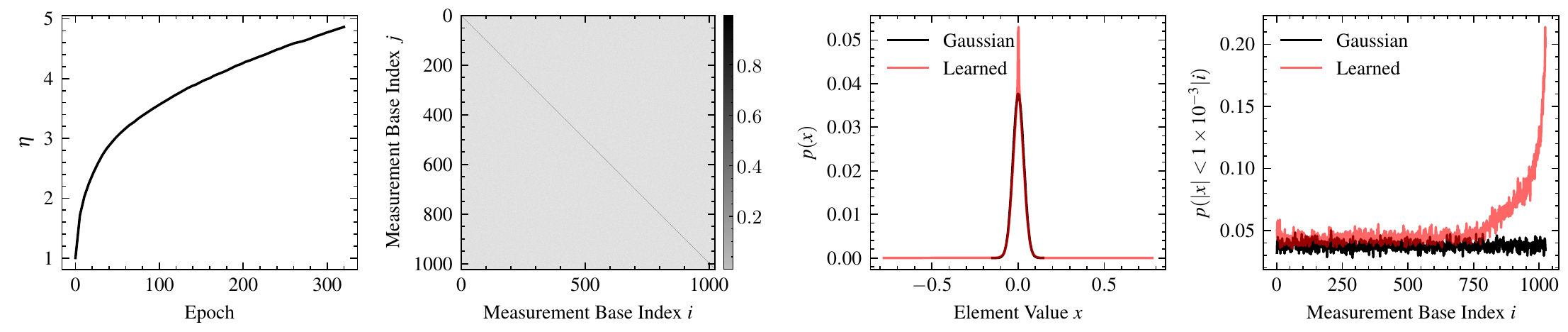}
		\vspace{-24pt}
		\caption{Illustrations of the logarithmic-like growth curve of $\eta$ (\textcolor{blue}{left}), visualization of the learned normalized $[(\widetilde{\mathbf{A}})(\widetilde{\mathbf{A}})^{\top}]$ with $\widetilde{\mathbf{A}}=\mathbf{A}/{\sqrt{\eta}}$ (\textcolor{blue}{middle left}), histograms of $\widetilde{\mathbf{A}}$ and fixed random Gaussian matrix (\textcolor{blue}{middle right}), and the proportion curves of measurement base elements close to zero (\textcolor{blue}{right}).}
		\label{fig: v_phi_dis}
		\vspace{-10pt}
	\end{figure*}

	\begin{table*}
		\centering
		\caption{Comparisons of the functional features, total parameter numbers for seven CS ratios and average inference time of sampling and recovering a $256\times 256$ image on an 1080Ti GPU with CS ratio $r=10\%$ among various network-based CS methods.}
		\vspace{-2pt}
		\resizebox{1.0\textwidth}{!}{%
			\label{tab:feature_comp}
			\begin{tabular}{|l|c|c|c|c|c|c|}
				\hline
				\multirow{2}{*}{Method}&Sampling Matrix&Adaptive CS&Fine Granular&Deblocking&CS Ratio Inform-&\#Param. (M)\\
				&Learnability&Ratio Allocation&Scalability&Ability&ation Exploitation&/Time (ms)\\
				\hline
				\hline
				ReconNet \cite{DL2_ReconNet} (CVPR 2016)&$\times$&$\times$&$\times$&$\times$&$\times$&0.98/2.69\\
				ISTA-Net$^+$ \cite{ISTA-Net} (CVPR 2018)&$\times$&$\times$&$\times$&$\times$&$\times$&2.38/5.65\\
				DPA-Net \cite{sun2020dual} (TIP 2020)&$\times$&$\times$&$\times$&$\checkmark$&$\times$&65.17/36.49\\
				ConvMMNet \cite{mdrafi2020joint} (TCI 2020)&$\checkmark$&$\times$&$\times$&$\times$&$\times$&7.64/19.32\\
				CSNet$^+$ \cite{DL3_CSNet} (TIP 2019)&$\checkmark$&$\times$&$\times$&$\checkmark$&$\times$&4.35/16.77\\
				OPINE-Net$^+$ \cite{OPINE-Net} (JSTSP 2020)&$\checkmark$&$\times$&$\times$&$\checkmark$&$\times$&4.35/17.31\\
				AMP-Net \cite{AMP-Net} (TIP 2021)&$\checkmark$&$\times$&$\times$&$\checkmark$&$\times$&6.08/27.38\\
				SCSNet \cite{SCSNet} (CVPR 2019)&$\checkmark$&$\times$&$\checkmark$&$\checkmark$&$\times$&0.80/30.91\\
				BCS-Net \cite{BCS-Net} (TMM 2020)&$\checkmark$&$\checkmark$&$\times$&$\checkmark$&$\times$&1.64/83.86\\
				COAST \cite{you2021coast} (TIP 2021)&$\checkmark$&$\times$&$\checkmark$&$\checkmark$&$\checkmark$&1.12/45.54\\ \cline{1-7}
				CASNet (Ours)&$\checkmark$&$\checkmark$&$\checkmark$&$\checkmark$&$\checkmark$&16.90/97.37\\
				\hline
		\end{tabular}}
		\vspace{-10pt}
	\end{table*}
	
	\begin{table*}
		\centering
		\caption{Average PSNR(dB)/SSIM performance comparisons among various CS methods on Set11 \cite{DL2_ReconNet} and CBSD68 \cite{CBSD68} with seven different CS ratios. The best and second best results are highlighted in {red} and {blue} colors, respectively.}
		\label{tab:comp}
		\vspace{-4pt}
		\resizebox{0.96\textwidth}{!}{%
			\begin{tabular}{|c|l|ccccccc|}
				\hline
				\multirow{2}{*}{Dataset}&\multirow{2}{*}{Method}&\multicolumn{7}{c|}{CS Ratio $r$}\\ \cline{3-9}
				&&1\%&4\%&10\%&25\%&30\%&40\%&50\%\\
				\hline
				\hline
				\multirow{10}{*}{Set11 \cite{DL2_ReconNet}}&ReconNet \cite{DL2_ReconNet}&17.43/0.4017&20.93/0.5897&24.38/0.7301&28.44/0.8531&29.09/0.8693&30.60/0.9020&32.25/0.9177\\
				&ISTA-Net$^{+}$ \cite{ISTA-Net}&17.48/0.4479&21.32/0.6037&26.64/0.8087&32.59/0.9254&33.68/0.9352&35.97/0.9544&38.11/0.9707\\
				&DPA-Net \cite{sun2020dual}&18.05/0.5011&23.50/0.7205&26.99/0.8354&31.74/0.9238&33.35/0.9425&35.21/0.9580&36.80/0.9685\\
				&ConvMMNet \cite{mdrafi2020joint}&19.53/0.4902&23.92/0.7384&27.63/0.8594&32.05/0.9300&33.25/0.9442&35.04/0.9579&36.72/0.9687\\
				&CSNet$^{+}$ \cite{DL3_CSNet}&20.67/0.5411&24.83/0.7480&28.34/0.8580&33.34/0.9387&34.27/0.9492&36.44/0.9690&38.47/0.9796\\
				&OPINE-Net$^{+}$ \cite{OPINE-Net}&20.15/0.5340&\textcolor{blue}{25.69}/\textcolor{blue}{0.7920}&29.81/0.8904&\textcolor{blue}{34.86}/\textcolor{blue}{0.9509}&35.79/0.9541&37.96/0.9633&40.19/0.9800\\
				&AMP-Net \cite{AMP-Net}&20.55/\textcolor{blue}{0.5638}&25.14/0.7701&29.42/0.8782&34.60/0.9469&35.91/0.9576&\textcolor{blue}{38.25}/\textcolor{blue}{0.9714}&40.26/0.9786\\
				&SCSNet \cite{SCSNet}&\textcolor{blue}{21.04}/0.5562&24.29/0.7589&28.52/0.8616&33.43/0.9373&34.64/0.9511&36.92/0.9666&39.01/0.9769\\ 
				&BCS-Net \cite{BCS-Net}&20.86/0.5510&24.90/0.7531&29.42/0.8673&34.20/0.9408&35.63/0.9495&36.68/0.9667&39.58/0.9734\\
				&COAST \cite{you2021coast}&$-$/$-$&$-$/$-$&\textcolor{blue}{30.03}/\textcolor{blue}{0.8946}&$-$/$-$&\textcolor{blue}{36.35}/\textcolor{blue}{0.9618}&$-$/$-$&\textcolor{blue}{40.32}/\textcolor{blue}{0.9804}\\ \cline{2-9}
				&CASNet (Ours)&\textcolor{red}{21.97}/\textcolor{red}{0.6140}&\textcolor{red}{26.41}/\textcolor{red}{0.8153}&\textcolor{red}{30.36}/\textcolor{red}{0.9014}&\textcolor{red}{35.67}/\textcolor{red}{0.9591}&\textcolor{red}{36.92}/\textcolor{red}{0.9662}&\textcolor{red}{39.04}/\textcolor{red}{0.9760}&\textcolor{red}{40.93}/\textcolor{red}{0.9826}\\
				\hline
				\hline
				\multirow{10}{*}{CBSD68 \cite{CBSD68}}&ReconNet \cite{DL2_ReconNet}&18.27/0.4007&21.66/0.5210&24.15/0.6715&26.04/0.7833&27.53/0.8045&29.08/0.8658&29.86/0.8951\\
				&ISTA-Net$^{+}$ \cite{ISTA-Net}&19.14/0.4158&22.17/0.5486&25.32/0.7022&29.36/0.8525&30.25/0.8781&32.30/0.9195&34.04/0.9424\\
				&DPA-Net \cite{sun2020dual}&20.25/0.4267&23.50/0.7205&25.47/0.7372&29.01/0.8595&29.73/0.8827&31.17/0.9156&32.55/0.9386\\
				&ConvMMNet \cite{mdrafi2020joint}&21.27/0.4805&24.21/0.6508&26.75/0.7831&30.16/0.8935&31.02/0.9134&32.81/0.9402&34.30/0.9570\\
				&CSNet$^{+}$ \cite{DL3_CSNet}&\textcolor{blue}{22.21}/0.5100&25.43/0.6706&27.91/0.7938&31.12/0.9060&32.20/0.9220&35.01/0.9258&36.76/0.9638\\
				&OPINE-Net$^{+}$ \cite{OPINE-Net}&22.11/0.5140&25.20/\textcolor{blue}{0.6825}&27.82/0.8045&\textcolor{blue}{31.51}/\textcolor{blue}{0.9061}&32.35/0.9215&34.95/0.9261&36.35/0.9660\\
				&AMP-Net \cite{AMP-Net}&22.18/\textcolor{blue}{0.5207}&\textcolor{blue}{25.47}/0.6534&27.79/0.7853&31.37/0.8749&32.68/0.9291&35.06/0.9395&36.59/0.9620\\
				&SCSNet \cite{SCSNet}&22.03/0.5126&25.37/0.6623&\textcolor{blue}{28.02}/0.8042&31.15/0.9058&32.64/0.9237&35.03/0.9214&36.27/0.9593\\
				&BCS-Net \cite{BCS-Net}&21.95/0.5119&25.44/0.6597&27.98/0.8015&31.29/0.8846&\textcolor{blue}{32.70}/\textcolor{blue}{0.9301}&\textcolor{blue}{35.14}/\textcolor{blue}{0.9397}&\textcolor{blue}{36.85}/\textcolor{blue}{0.9682}\\
				&COAST
				\cite{you2021coast}&$-$/$-$&$-$/$-$&27.92/\textcolor{blue}{0.8061}&$-$/$-$&32.66/0.9256&$-$/$-$&36.43/0.9663\\ \cline{2-9}
				&CASNet (Ours)&\textcolor{red}{22.49}/\textcolor{red}{0.5520}&\textcolor{red}{25.73}/\textcolor{red}{0.7079}&\textcolor{red}{28.41}/\textcolor{red}{0.8231}&\textcolor{red}{32.31}/\textcolor{red}{0.9196}&\textcolor{red}{33.40}/\textcolor{red}{0.9359}&\textcolor{red}{35.43}/\textcolor{red}{0.9581}&\textcolor{red}{37.48}/\textcolor{red}{0.9728}\\
				\hline
		\end{tabular}}
		\vspace{-14pt}
	\end{table*}

	\begin{figure*}[t]
		\centering
		\includegraphics[width=0.9\textwidth]{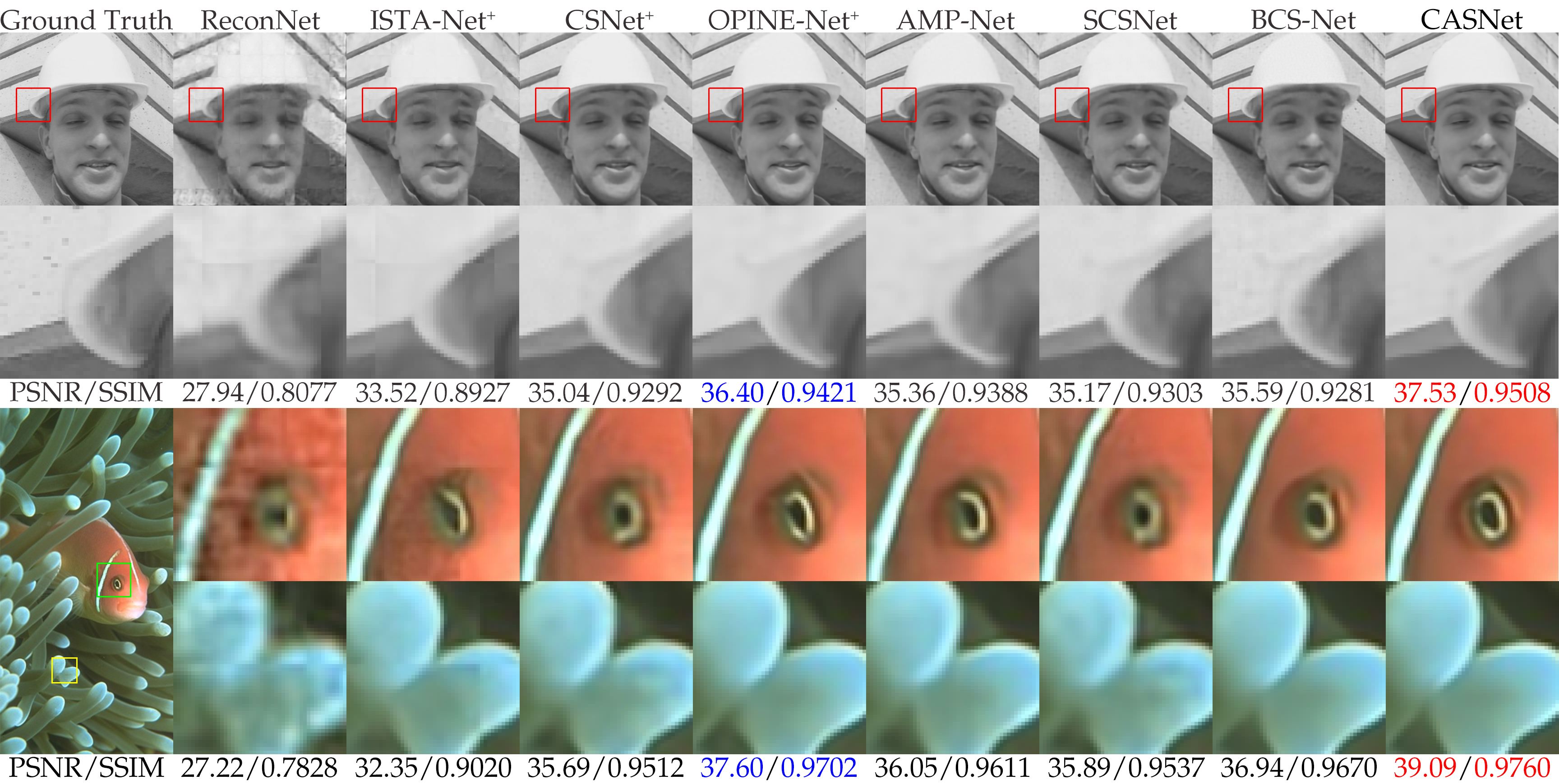} 
		\vspace{-8pt}
		\caption{Visual comparisons with PSNR(dB)/SSIM on recovering two images from Set11 \cite{DL2_ReconNet} (\textcolor{blue}{top}) and CBSD68 \cite{CBSD68} (\textcolor{blue}{bottom}) respectively in the case of CS ratio $r=10\%$. Our CASNet shows its superiority by giving better visual results with more details and shaper edges.}
		\label{fig: v_comp_set11}
		\vspace{-14pt}
	\end{figure*}

	\begin{figure}[!t]
		\begin{center}
			\scalebox{0.57}{
				\begin{tabular}[b]{c@{ } c@{ }  c@{ } c@{ } c@{ }}\hspace{-4mm}
					\multirow{4}{*}{\includegraphics[width=.307\textwidth,valign=t]{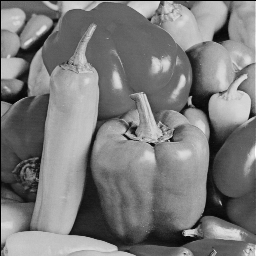}} &   
					\includegraphics[trim={3cm 3cm  3cm  3cm },clip,width=.13\textwidth,valign=t]{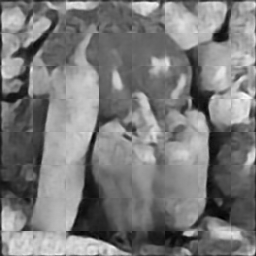}&
					\includegraphics[trim={3cm 3cm 3cm 3cm },clip,width=.13\textwidth,valign=t]{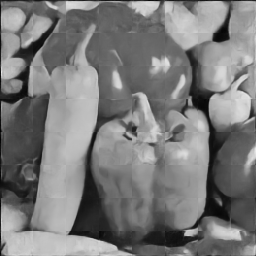}&   
					\includegraphics[trim={3cm 3cm 3cm 3cm },clip,width=.13\textwidth,valign=t]{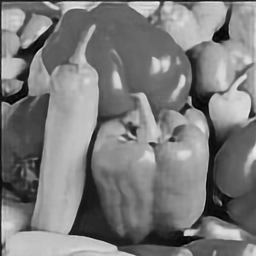}&
					\includegraphics[trim={3cm 3cm 3cm 3cm },clip,width=.13\textwidth,valign=t]{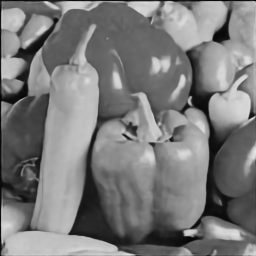}
					
					\\
					&  \small~23.22/0.6914 &\small~27.12/0.8151  & \small~27.85/0.8824& \small~\textcolor{blue}{30.33}/\textcolor{blue}{0.9094}\\
					& \small~ReconNet& \small~ISTA-Net$^{+}$& \small~CSNet$^{+}$& \small~OPINE-Net$^{+}$\\
					
					&
					\includegraphics[trim={3cm 3cm 3cm 3cm },clip,width=.13\textwidth,valign=t]{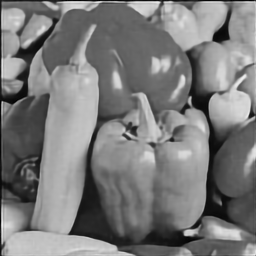}&
					\includegraphics[trim={3cm 3cm 3cm 3cm },clip,width=.13\textwidth,valign=t]{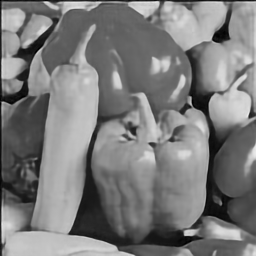}&
					\includegraphics[trim={3cm 3cm 3cm 3cm },clip,width=.13\textwidth,valign=t]{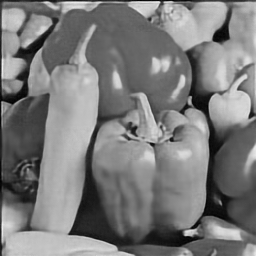}&  
					\includegraphics[trim={3cm 3cm 3cm 3cm },clip,width=.13\textwidth,valign=t]{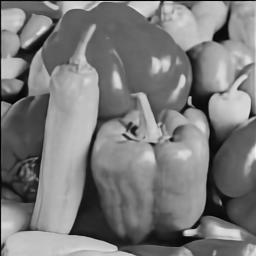}\\
					
					\small~PSNR/SSIM& \small~29.88/0.9079& \small~28.22/0.8874& \small~28.48/0.8895
					& \small~\textcolor{red}{31.60}/\textcolor{red}{0.9240}\\
					\small~Ground Truth& \small~AMP-Net& \small~SCSNet& \small~BCS-Net& \small~CASNet \footnotesize(Ours)
					\\
			\end{tabular}}
			\scalebox{0.57}{
				\begin{tabular}[b]{c@{ } c@{ }  c@{ } c@{ } c@{ }}\hspace{-4mm}
					\multirow{4}{*}{\includegraphics[trim={96 0  96  0 },clip,width=.307\textwidth,valign=t]{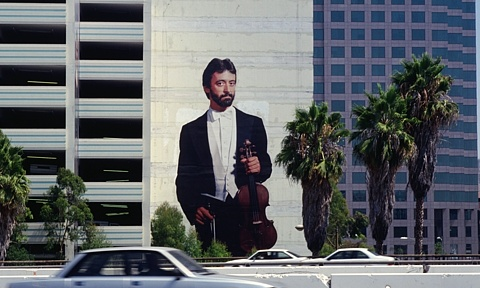}} &   
					\includegraphics[trim={180 160  220  48 },clip,width=.13\textwidth,valign=t]{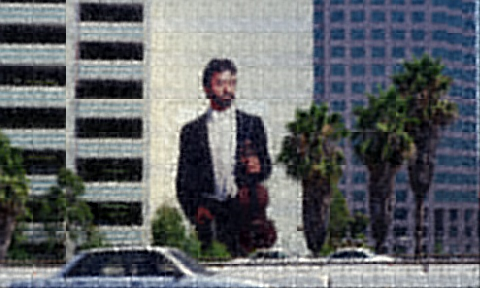}&
					\includegraphics[trim={180 160  220  48 },clip,width=.13\textwidth,valign=t]{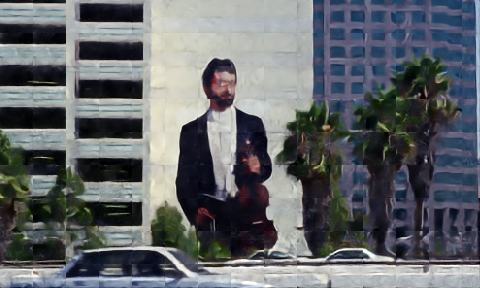}&   
					\includegraphics[trim={180 160  220  48 },clip,width=.13\textwidth,valign=t]{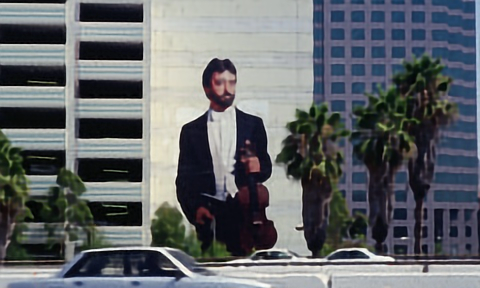}&
					\includegraphics[trim={180 160  220  48 },clip,width=.13\textwidth,valign=t]{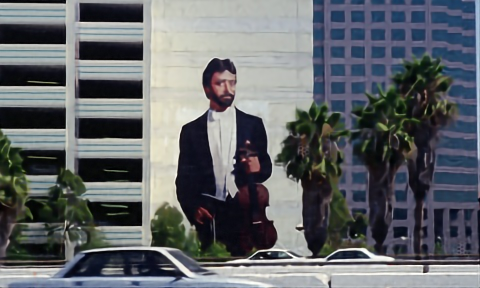}
					
					\\
					&  \small~21.35/0.5973 &\small~23.09/0.6859  & \small~26.59/0.8161& \small~26.50/0.8416\\
					& \small~ReconNet& \small~ISTA-Net$^{+}$& \small~CSNet$^{+}$& \small~OPINE-Net$^{+}$\\
					
					&
					\includegraphics[trim={180 160 220  48 },clip,width=.13\textwidth,valign=t]{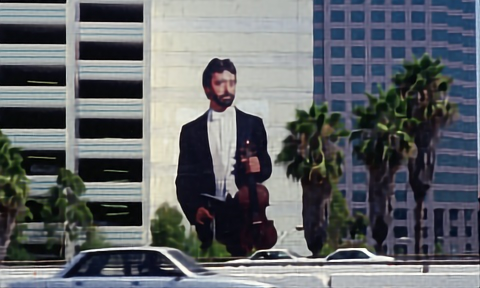}&
					\includegraphics[trim={180 160 220  48 },clip,width=.13\textwidth,valign=t]{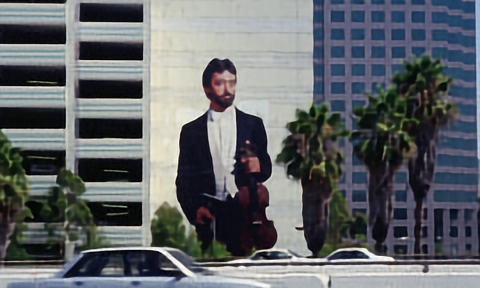}&
					\includegraphics[trim={180 160 220  48 },clip,width=.13\textwidth,valign=t]{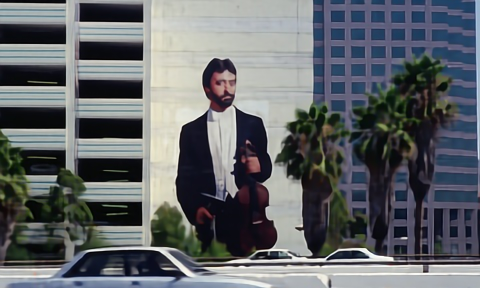}&  
					\includegraphics[trim={180 160 220  48 },clip,width=.13\textwidth,valign=t]{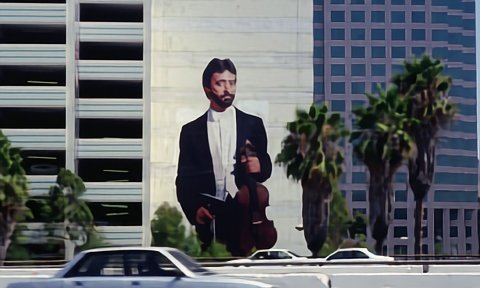}\\
					
					\small~PSNR/SSIM& \small~26.48/\textcolor{blue}{0.8510}& \small~\textcolor{blue}{26.71}/0.8188& \small~26.38/0.8473
					& \small~\textcolor{red}{27.35}/\textcolor{red}{0.8744}\\
					\small~Ground Truth& \small~AMP-Net& \small~SCSNet& \small~BCS-Net& \small~CASNet \footnotesize(Ours)
					\\
			\end{tabular}}
			\scalebox{0.57}{
				\begin{tabular}[b]{c@{ } c@{ }  c@{ } c@{ } c@{ }}\hspace{-4mm}
					\multirow{4}{*}{\includegraphics[trim={160 0  32  0 },clip,width=.307\textwidth,valign=t]{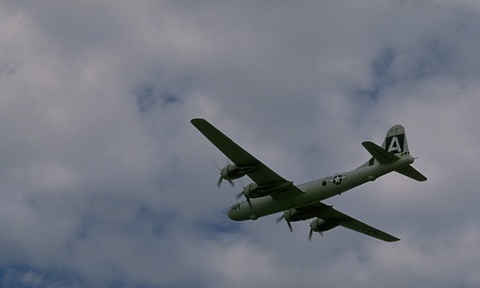}} &   
					\includegraphics[trim={180 100  220  108 },clip,width=.13\textwidth,valign=t]{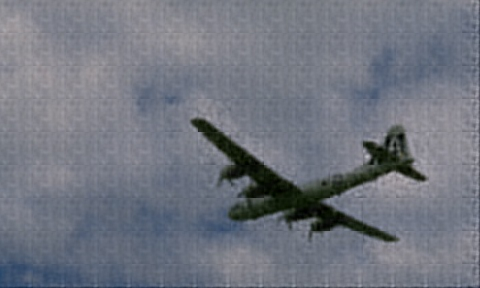}&
					\includegraphics[trim={180 100  220  108 },clip,width=.13\textwidth,valign=t]{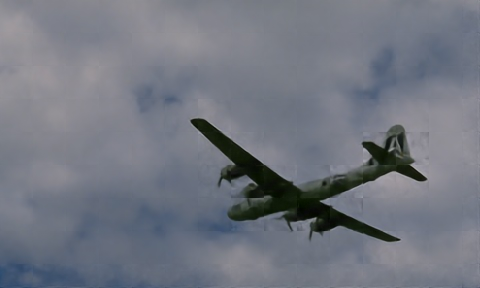}&   
					\includegraphics[trim={180 100  220  108 },clip,width=.13\textwidth,valign=t]{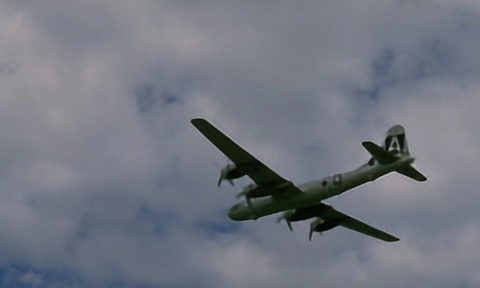}&
					\includegraphics[trim={180 100  220  108 },clip,width=.13\textwidth,valign=t]{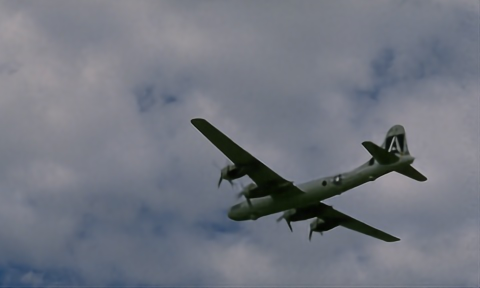}
					
					\\
					&  \small~30.66/0.7816 &\small~37.07/0.9618 & \small~\textcolor{blue}{40.89}/0.9809& \small~\textcolor{blue}{40.89}/0.9807\\
					& \small~ReconNet& \small~ISTA-Net$^{+}$& \small~CSNet$^{+}$& \small~OPINE-Net$^{+}$\\
					
					&
					\includegraphics[trim={180 100  220  108 },clip,width=.13\textwidth,valign=t]{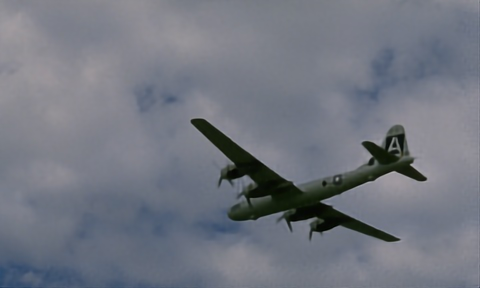}&
					\includegraphics[trim={180 100  220  108 },clip,width=.13\textwidth,valign=t]{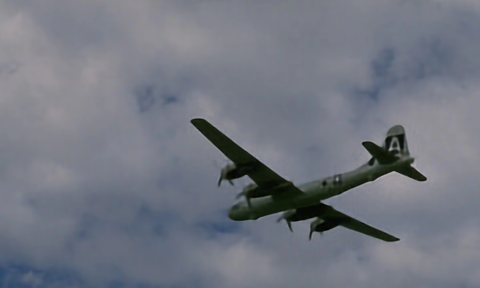}&
					\includegraphics[trim={180 100  220  108 },clip,width=.13\textwidth,valign=t]{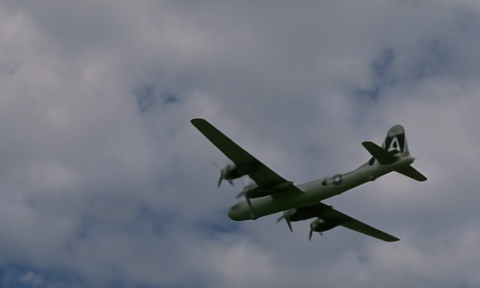}&  
					\includegraphics[trim={180 100  220  108 },clip,width=.13\textwidth,valign=t]{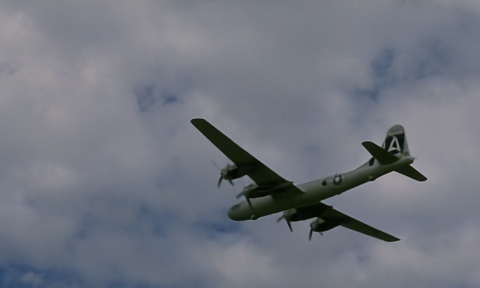}\\
					
					\small~PSNR/SSIM& \small~40.17/0.9789& \small~40.86/0.9805& \small~40.49/\textcolor{blue}{0.9822}
					& \small~\textcolor{red}{41.78}/\textcolor{red}{0.9847}\\
					\small~Ground Truth& \small~AMP-Net& \small~SCSNet& \small~BCS-Net& \small~CASNet \footnotesize(Ours)
					\\
			\end{tabular}}
		\end{center}
		\vspace{-12pt}
		\caption{Visual comparisons with PSNR(dB)/SSIM on recovering an image from Set11 \cite{DL2_ReconNet} (\textcolor{blue}{top}) and two images from CBSD68 \cite{CBSD68} (\textcolor{blue}{middle} and \textcolor{blue}{bottom}, respectively) in the case of CS ratio $r=10\%$.}
		\label{fig: vis_comp_more}
		\vspace{4pt}
	\end{figure}

	\begin{figure}[t]
		\centering
		\includegraphics[width=0.48\textwidth]{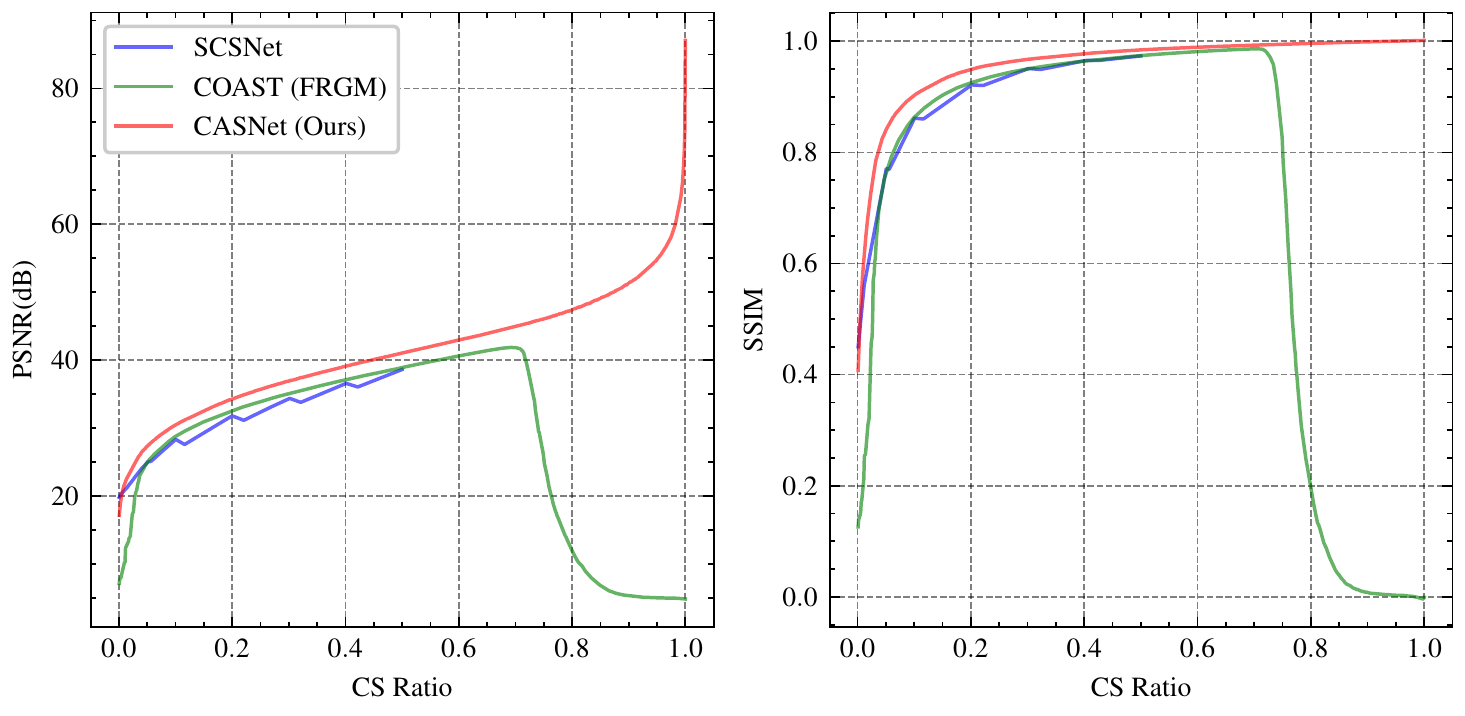}
		\vspace{-8pt}
		\caption{Comparison of fine granular scalable CS reconstruction performance among SCSNet \cite{SCSNet}, the default COAST version with fixed random Gaussian matrices (FRGM) \cite{you2021coast} and our CASNet on Set11 \cite{DL2_ReconNet}.}
		\label{fig: scalable_curve}
		\vspace{-10pt}
	\end{figure}
	
	\begin{figure}
		\hspace{8pt}
		\includegraphics[width=0.45\textwidth]{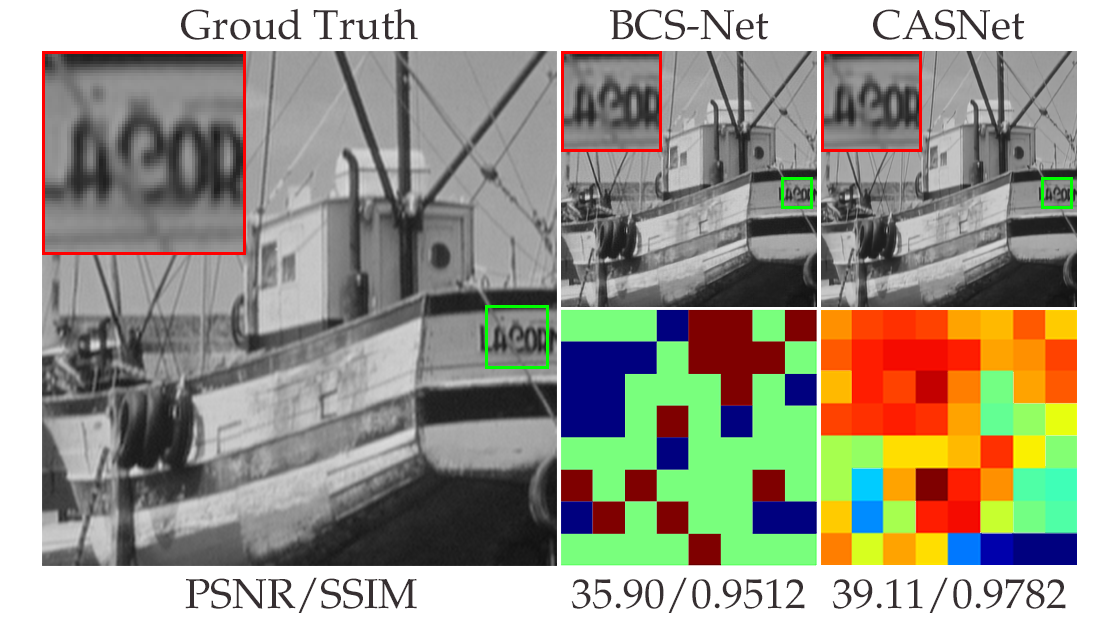}
		\vspace{-10pt}
		\caption{Comparion between two saliency-based method on the image named ``Boats" (\textcolor{blue}{left}) from Set11 \cite{DL2_ReconNet} with CS ratio $r=30\%$. The recovered results are in \textcolor{blue}{top middle} and \textcolor{blue}{top right}. In addition, here we provide their CS ratio allocating results in \textcolor{blue}{bottom middle} and \textcolor{blue}{bottom right}.}
		\label{fig: bcs}
		\vspace{4pt}
	\end{figure}

	\textbf{(6) Study of Phase Number $N_p$:} Under the PGD-unfolding framework, our CASNet is expected to get higher performance with RS phase number $N_p$ increases. Fig. \ref{fig: phase_num} gives the comparison among CASNet variants with different phase numbers with CS ratio $r=25\%$. We can see that both the curves increase as $N_p$ increases, but they are becoming almost flat when $N_p\ge 13$. Therefore, we choose the CASNet with 13 phases as the default setting by considering the trade-off between model complexity and recovery accuracy.
	
	\textbf{(7) Analysis of the Learned $\mathbf{A}$:} To get the insight of sampling matrix structure, we give some findings of the learned generating matrix $\mathbf{A}$. First, although there is no constraints on $\mathbf{A}$ in loss function, the orthogonal form $\mathbf{A}\mathbf{A}^{\top}=\eta \mathbf{I}$ is approximately satisfied during training, where $\eta$ shows a logarithmic-like growth trend from 1 to about 4.865 as illustrated in the \textcolor{blue}{left} subgraph of Fig. \ref{fig: v_phi_dis}. The \textcolor{blue}{middle left} subgraph visualizes $[(\widetilde{\mathbf{A}})(\widetilde{\mathbf{A}})^{\top}]$ with the normalized $\widetilde{\mathbf{A}}={\mathbf{A}}/{\sqrt{\eta}}$, in which the position $(u, v)$ corresponds to the inner product of the $u$-th and the $v$-th normalized bases. We observe that all diagonal elements are about 1.0 and others are near to 0. Second, we plot the histograms and the proportion of base elements close to zero of $\widetilde{\mathbf{A}}$ and the fixed random one as shown in the \textcolor{blue}{right} two subgraphs. We observe that the learned one exhibits wider and sparser distribution, and the base sparsity gets higher as the index increases. Third, we reshape the first eight learned rows into $B\times B$ and visualize them in Fig. \ref{fig: v_phi} with frequency. They show structured and anisotropic spatial results different from traditional manually defined filters. And frequencies of the former bases are narrower, which means that they pay more attention to low-frequency information. These facts verify our ${\mathbf{A}}$ design with a descending base importance order and the feasibility of its data-driven learning scheme.
	
	\vspace{-8pt}
	\subsection{Comparison with State-of-the-Arts}
	\vspace{-2pt}
	We compare CASNet with ten representative state-of-the-art CS networks: ReconNet \cite{DL2_ReconNet}, ISTA-Net$^{+}$ \cite{ISTA-Net}, CSNet$^{+}$ \cite{DL3_CSNet}, DPA-Net \cite{sun2020dual}, ConvMMNet \cite{mdrafi2020joint}, OPINE-Net$^{+}$ \cite{OPINE-Net}, AMP-Net \cite{AMP-Net}, SCSNet \cite{SCSNet} BCS-Net \cite{BCS-Net} and COAST \cite{you2021coast}. ReconNet, CSNet$^{+}$, DPA-Net and ConvMMNet are traditional network-based methods; ISTA-Net$^{+}$, OPINE-Net$^{+}$ and AMP-Net are traditional unfolding methods; SCSNet is scalable with a hierarchical structure; BCS-Net is saliency-based and achieves CS ratio allocation with a multi-channel architecture; COAST achieves scalability by generalizing to arbitrary sampling matrices. More details of high-level functional feature comparisons are given in Tab. \ref{tab:feature_comp}, where CASNet exhibits its organic integration of different features and merits.
	
	\textbf{(1) Average PSNR/SSIM Comparisons on Benchmarks:} The PSNR/SSIM results of different CS methods on Set11 \cite{DL2_ReconNet} and CBSD68 \cite{CBSD68} for seven CS ratios are provided in Tab. \ref{tab:comp}. Despite the fact that CSNet$^{+}$, OPINE-Net$^{+}$ and AMP-Net learn separate models and exceed the first three based on fixed random sampling matrices, they still fail to outperform CASNet and remain the average PSNR/SSIM distances of 1.38dB/0.0252 and 0.65dB/0.0253 on Set11 and CBSD68, respectively. Comparisons with SCSNet, BCS-Net and COAST (the default version with sampling matrices from learned OPINE-Nets) show the superiority of CASNet which combines our three main ideas with a well-defined structure and boosting schemes. Note that the uniform samping CASNet version (see Tab. \ref{tab:ablation} (1)) is already able to show a large PSNR exceeding over 1dB on Set11 \cite{DL2_ReconNet} compared with the existing best methods, and our saliency-based allocation can further breakthrough the accuracy saturation and bring a large step forward on this basic variant. The visual comparisons in Fig. \ref{fig: v_comp_set11} and Fig. \ref{fig: vis_comp_more} show that our CASNet is able to recover high-quality results with more details and sharper edges.
	
	\textbf{(2) Comparisons of Scalable Sampling and Recovery:} The recovery accuracy curves in Fig. \ref{fig: scalable_curve} illustrate that CASNet outperforms the scalable SCSNet \cite{SCSNet} and COAST \cite{you2021coast} with a large margin in nearly all cases. Here we choose the default COAST version with fixed random Gaussian matrices (FRGM) as it is expensive to train thousands of OPINE-Nets with different CS ratios and utilize the learned matrices as described in \cite{you2021coast}. Compared with using a hierarchical CNN and a greedy method to get the base importance order to support multiple CS ratios in $\left[0, 0.5\right]$ by SCSNet, and adopting a random projection augmentation strategy with CS ratios in $\left\{ 0.1,0.2,0.3,0.4,0.5 \right\}$ for training by COAST which works normally in $\left[0.05, 0.7\right]$ but gets failed in other cases, our scalable scheme based on $\mathbf{A}$ achieves the more robust fine granular scalability with the complete CS ratio range of $\left[0,1\right]$, provides a much easier and intuitive joint learning method, and brings us a more clear framework with strong interpretability.
	
	\textbf{(3) Comparisons of Saliency-based CS Ratio Allocation:} The recovery and CS ratio allocation results by BCS-Net \cite{BCS-Net} and CASNet in Fig. \ref{fig: bcs} shows the flexibility of CASNet assigning a variety of CS ratio levels with smooth transition among blocks and its powerful recovery ability with a PSNR/SSIM exceeding of 3.21dB/0.0270 compared with BCS-Net, which adopts a seven-channel sampling method based on a handcrafted saliency detecting method and only assigns three ratio levels to the blocks.
	
	\vspace{-3pt}
	\section{Conclusion and Future Work}
	A novel content-aware scalable network named CASNet is proposed to comprehensively address image CS problems, which tries to make full use of the merits of traditional methods by achieving adaptive CS ratio allocation, fine granular scalability, and high-quality reconstruction collectively. Different from the previous saliency-based methods, we use a data-driven saliency detector and a block ratio aggregation (BRA) strategy to achieve accurate sampling rate allocations. A unified learnable generating matrix is developed to produce sampling matrices with memory complexity reduction. The PGD-unfolding recovery subnet exploits the CS ratio information and the inter-block relationship to restore images step-by-step. We use an SVD-based initialization scheme to accelerate training, and a random transformation enhancement (RTE) strategy to improve the network robustness. All the CASNet parameters can be indiscriminately learned end-to-end with strong compatibility and mutual supports among its components and strategies. Furthermore, we consider the possible gap between the CASNet framework and physical CS systems, and provide a four-stage implementation for fair evaluations and practical deployments. Extensive experiments demonstrate that CASNet greatly improves upon the results of state-of-the-art CS methods with high structural efficiency and deep matrix insights. Our future work is to extend CASNet to video CS problems, and Fourier-based medical applications \cite{chun2017compressed} like CS-MRI \cite{MRI1, MRI2} and sparse-view CT \cite{szczykutowicz2010dual} tasks.

	\begin{figure}[!h]
            \vspace{-16pt}
		\centering
		\hspace{-10pt}
		\includegraphics[width=0.5065\textwidth]{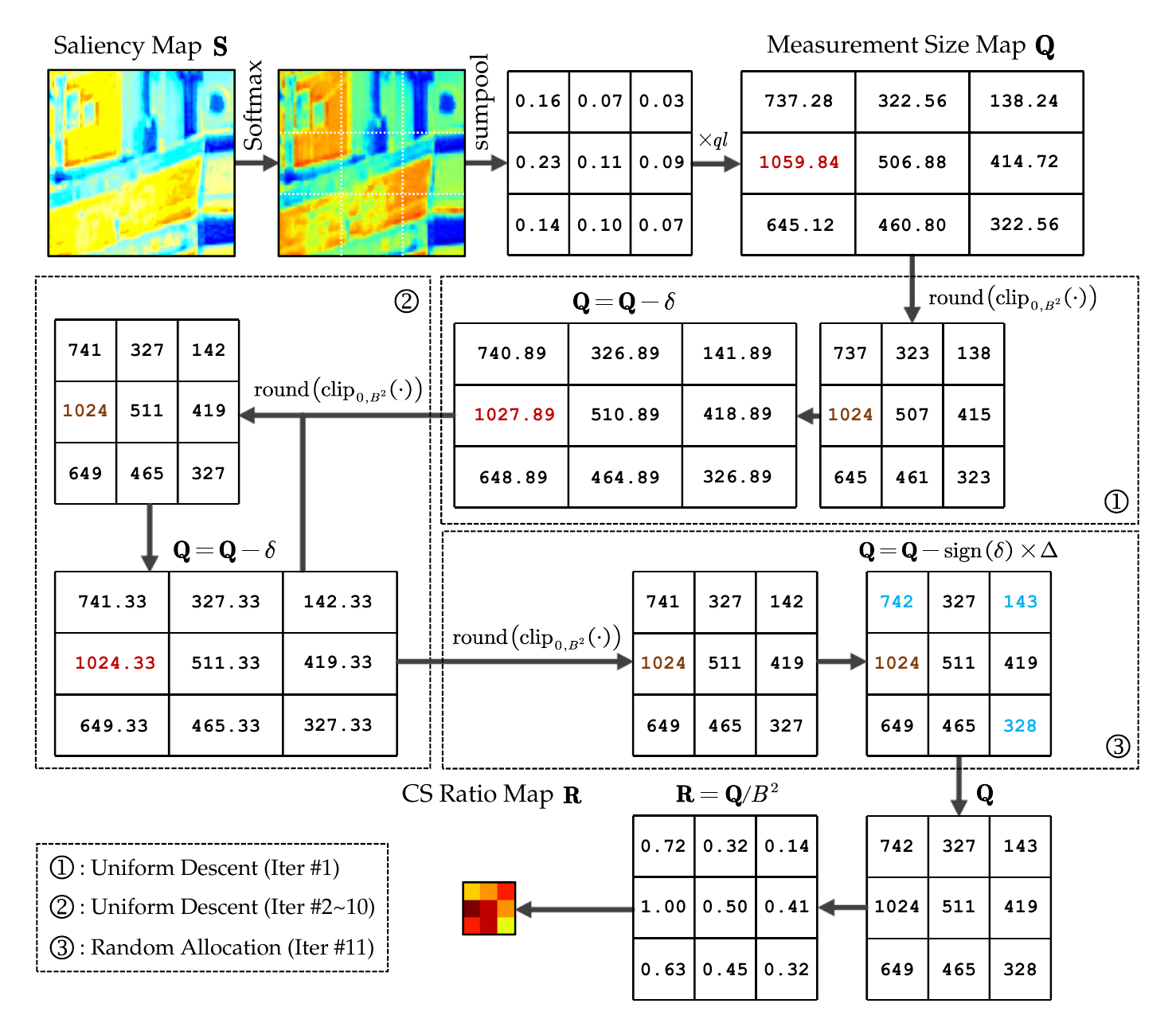}
		\vspace{-15pt}
		\caption{Illustration of a simple but complete process of an accurate and fast CS ratio allocation performed by our proposed BRA strategy, which distributes the allocated CS ratio map $\mathbf{R}$ with 11 iterations in the case of $B=32$, $l=9$, target CS ratio $r=50\%$ ($q=512$) and $K=N=1024$. As we can see, the BRA strategy, which consists of softmax normalization, sumpooling aggregation, and error correction, takes $\{\mathbf{S},B,q,K\}$ as inputs and gives out the accurately allocated CS ratio map $\mathbf{R}$ to conduct the content-aware samplings of CASNet framework and its practical implementations.}
		\label{fig: instance_BRA}
		\vspace{6pt}
	\end{figure}

	\vspace{-3pt}
	\appendices
	\section{A Simple Instance of BRA Strategy}
	\label{instance_BRA}
	In order to provide a clear exhibition and a better understanding of our BRA strategy, here we provide an instance of processing the saliency map $\mathbf{S}$ of a $96\times 96$ image patch and give out the allocated CS ratio map $\mathbf{R}$ with a practical configuration of CS ratio $r=50\%$, \textit{i.e.}, $B=32$, $l=9$, $q=512$ and $K=N=1024$. As illustrated in Fig. \ref{fig: instance_BRA}, $\mathbf{S}$ is first normalized by the softmax operator in spatial to obtain the weight map, and initially aggregated by the $32\times 32$ sumpooling to get the weights of all blocks. The aggregated weight map is then timed by the target sum of block measurement sizes $512\times 9$ to obtain the measurement size map $\mathbf{Q}$. Since the measurement sizes of blocks can only be integers in the range of $\left[0, 1024\right]$ and their sum should be $512\times 9$ for accuracy, the initial $\mathbf{Q}$ is sent to acquire corrections iteratively. In each iteration, we first use $\text{round}(\text{clip}_{0,1024}(\cdot))$ to shear and discretize the allocated measurement size of each block independently, then the average error $\delta =\text{average}(\mathbf{Q}) - 512$ determines whether the correction should be performed. Once $\delta$ equals zero, then the error correction stage can be stopped and $\mathbf{Q}$ is finally normalized by $\mathbf{R}={\mathbf{Q}}/{1024}$ to get the CS ratio map $\mathbf{R}$. Otherwise, $\mathbf{Q}$ needs to be further corrected by our provided two correcting methods. The first method is uniform descent, which directly employs $\mathbf{Q}=\mathbf{Q}-\delta$ to make a fair adjustment. However, adopting the uniform descent only could result in a dead loop in some cases since the remained error may not be eliminated (quantitative analysis is provided in Appx. \ref{convergence_analysis_BRA}), so we provide the second correcting method to randomly distribute the residual distances to blocks based on the multinomial distribution. In our default setting, the uniform descent has a maximum iteration number $T=10$, which means that if the first method can not eliminate the error in the first ten iterations, the random approach will be utilized for further corrections. The instance in Fig. \ref{fig: instance_BRA} gives an example of the BRA work principle, which exhibits the details of our two correcting methods for the initial measurement size map.

	
	\section{Convergence Analysis of BRA Strategy}
	\label{convergence_analysis_BRA}
	
	Here we provide convergence analysis of BRA strategy in Alg. \ref{BRA} to demonstrate its effectiveness. We first show that the first part of BRA: the uniform descent (marked by \textcolor{blue}{method \#1}) makes CS measurement size map $\mathbf{Q}$ convergent to a fixed point $\mathbf{Q}_\infty$ with a bounded average error $\delta_\infty\in(-1/2,1/2]$.
	
	We denote $t$ as the decent step index, $\mathbf{Q}_t$ and $\delta_t$ as the measurement size map and the average error in the $t$-th correction iteration (corresponding to the 7-th and 8-th numbered lines of Alg. \ref{BRA}), respectively, and $q_t^i$ as the $i$-th elememt of $\mathbf{Q}_t$. Given any initial measurement size map $\mathbf{Q}_0$ (corresponding to the 3-rd numbered line), the parameter set $\{l,q,K\}$ (block number, target average measurement size and upper bound), we denote the discretization function $f:\mathbb{R}\mapsto \{0, 1, \cdots,  K\}$ with $f(\cdot)=\text{round}(\text{clip}_{0,K}(\cdot))$ and provide four lemmas for facilitating the convergence analysis behind them as follows:
	
	\vspace{4pt}
	\noindent \textbf{\textcolor{blue}{Lemma 1} (``Shape convergence" of $\mathbf{Q}$).} \textit{For series $\{\mathbf{Q}_t\}_{t=0}^T$, $\forall i\ne j$, it holds that: \textcolor{blue}{\textbf{(1)}} the larger (or equal) elements in $\mathbf{Q}_t$ remain larger (or equal) in $\mathbf{Q}_{t+1}$, i.e, $(q_t^i\ge q_t^j)\Rightarrow (q_{t+1}^i\ge q_{t+1}^j)$, and \textcolor{blue}{\textbf{(2)}} absolute difference between elements is monotonically decreasing, i.e., $|q_t^i-q_t^j|\ge |q_{t+1}^i-q_{t+1}^j|$.}
	
	\vspace{4pt}
	\noindent \textbf{\textcolor{purple}{Proof}.} \textit{For \textcolor{blue}{\textbf{(1)}}, since $\forall 1\le i\le l$, $q_{t+1}^i=f(q_{t}^i-\delta_t)$ and $\forall \delta_t\in\mathbb{R}$, $f((\cdot)-\delta_t)$ is monotonically increasing, the order-preservation among map elements holds. For \textcolor{blue}{\textbf{(2)}}, it is obvious that $\forall s\in\mathbb{R}$, $f(s+1)-f(s)\le 1$ (proof is omitted), we have:}
	\textit{\begin{align*}
	[-f(s)\le 1-f(s+1)]\Leftrightarrow [s-f(s)\le (s+1)-f(s+1)].
	\end{align*}}\textit{Let $s=x-\delta_t$, then it holds that:}
	\textit{\begin{align*}&[x-\delta_t-f(x-\delta_t)\le (x-\delta_t+1)-f(x-\delta_t+1)]\\\Leftrightarrow ~&[x-f(x-\delta_t)\le (x+1)-f(x+1-\delta_t)],\end{align*}}\textit{i.e., $[(\cdot)-f((\cdot)-\delta_t)]$ is increasing on $\{0,1,\cdots,K\}$. Assume that $q_t^i\ge q_t^j$, we have $q_t^i-f(q_t^i-\delta_t)\ge q_t^j-f(q_t^j-\delta_t)$, i.e., $q_t^i-q_t^j\ge q_{t+1}^i-q_{t+1}^j$. Similarly, when $q_t^i\le q_t^j$, it holds that $q_t^j-q_t^i\ge q_{t+1}^j-q_{t+1}^i$, i.e., $|q_t^i-q_t^j|\ge |q_{t+1}^i-q_{t+1}^j|$. $\blacksquare$}
	
	\vspace{4pt}
	The \textbf{Lemma 1} indicates that differences between each two $\mathbf{Q}$ elements converge to fixed values, and the whole map may then shift up and down by constants across all elements. Next, we show that $\{\mathbf{Q_t}\}$ will converge and not oscillate like that.
	
	\vspace{4pt}
	\noindent \textbf{\textcolor{blue}{Lemma 2} (Bound for absolute error).} \textit{$\forall n\in\{0,1,\cdots,K\}$ and $\delta\in \mathbb{R}$, it holds that: $|n-f(n-\delta)|\le\text{round}(\text{abs}(\delta))$.}
	
	\vspace{4pt}
	\noindent \textbf{\textcolor{purple}{Proof}.} \textit{Let $\text{floor}(x)$ and $\text{frac}(x)$ be the decimal and fractional parts of $x\in[0,+\infty)$, respectively. When $\delta\ge 0$, it holds that:}\textit{\begin{align*}&|n-f(n-\delta)|=n-f\left( n-\text{floor}\left( \delta \right) -\text{frac}\left( \delta \right) \right)\\=&\left\{ \begin{array}{l}
	n,\ n-\text{floor}\left( \delta \right) \le 0\\\text{round}\left(  \delta  \right) ,\ [0<n-floor\left( \delta \right) \le K] \land [\text{frac}(\delta)\ne 1/2]\\\text{floor}\left(  \delta  \right) ,\ [0<n-floor\left( \delta \right) \le K] \land [\text{frac}(\delta)= 1/2]	\end{array}\right.,\end{align*}}\textit{and when $\delta\le 0$, it holds that:}\textit{\begin{align*}|n-f(n-\delta)|=&f\left( n+\text{floor}\left(-\delta \right)+\text{frac}\left(-\delta \right) \right)-n\\=&\left\{ \begin{array}{l}\text{round}\left( -\delta \right) ,\ 0\le n+floor\left( -\delta \right) <K\\
	K-n,\ n+\text{floor}\left( -\delta \right) \ge K\\\end{array} \right..\end{align*}}\textit{For all cases, it holds that $|n-f(n-\delta)|\le\text{round}(\text{abs}(\delta))$. $\blacksquare$}
	
	\vspace{4pt}
	\noindent \textbf{\textcolor{blue}{Lemma 3} (Upper bound for $\text{round}(\cdot)$).} \textit{$\forall \delta\ge 0$, it holds that: $\text{round}(\delta)\le 2\delta$ and $[\text{round}(\delta)= 2\delta]\Leftrightarrow [\delta\in\{0,1/2\}]$.}
	
	\vspace{4pt}
	\noindent \textbf{\textcolor{purple}{Proof}.} \textit{When $\delta=0$, we have: $\text{round}(\delta)=0=2\delta$. When $0< \delta <1/2$, we have: $\text{round}(\delta)=0<2\delta$. When $\delta=1/2$, we have: $\text{round}(\delta)=1=2\delta$. When $1/2<\delta\le 1$, we have: $\text{round}(\delta)=1<2\delta$. When $1<\delta< 2$, we have: $\text{round}(\delta)\le\text{floor}(\delta)+1<2\text{floor}(\delta)+2\text{frac}(\delta)=2\delta$. And when $\delta\ge 2$, we have: $\text{round}(\delta)\le\text{floor}(\delta)+1<2\text{floor}(\delta)\le 2\delta$. $\blacksquare$}
	
	\vspace{4pt}
	Based on the above three lemmas, we present the following lemma and theorem to show convergency of uniform descent:
	
	\vspace{4pt}
	\noindent \textbf{\textcolor{blue}{Lemma 4} (Convergence of the absolute average error).} \textit{The absolute average error series $\{|\delta_t|\}_{t=0}^T$ converges as $T\rightarrow \infty$.}
	
	\vspace{4pt}
 	\noindent \textbf{\textcolor{purple}{Proof}.} \textit{When $\delta_t =0$, for a specific $t>0$, $f(\cdot)$ will not change any element value of $\mathbf{Q}_t$ and the latter $\delta$s will be all zeros (converged). And when $\delta_t\ne 0$, the ratio of $|\delta_{t+1}|$ to $|\delta_{t}|$ is:}\textit{\begin{align*}\left| \frac{\delta _{t+1}}{\delta _t} \right|=&\left| \frac{\frac{1}{l}\sum_{i=1}^l{[f\left( q_t^i-\delta _t \right) -q]}}{\delta _t} \right|\\=&\left| \frac{\frac{1}{l}\sum_{i=1}^l{[f\left( q_t^i-\delta _t \right) -q_t^i+q_t^i-q]}}{\frac{1}{l}\sum_{i=1}^l{(q_t^i-q)}} \right|\\=&\left| 1-\frac{d_t}{\delta _t} \right|,\ \text{where\ }d_t=\frac{1}{l}\sum_{i=1}^l{[q_{t}^{i}-f\left( q_{t}^{i}-\delta _t \right)]}.\end{align*}}\textit{Using \textbf{Lemma 2}, when $\delta_t >0$, we have: $d_t\le\text{round}(\delta_t)$, and when $\delta_t <0$, we have: $d_t\ge-\text{round}(-\delta_t)$. Using \textbf{Lemma 3}, we have: $0\le d_t/\delta_t \le 2$ and $|\delta_{t+1}/\delta_t|\le 1$. $\blacksquare$}
 	
 	\vspace{4pt}
	\noindent \textbf{\textcolor{red}{Theorem 1} (Convergence of measurement size map in the uniform descent).} \textit{The series $\{\mathbf{Q}_t\}_{t=0}^T$ converges to a fixed point $\mathbf{Q}_\infty$ with average error $\delta_\infty\in(-1/2,1/2]$ as $T\rightarrow \infty$.}
	
	\vspace{4pt}
 	\noindent \textbf{\textcolor{purple}{Proof}.} \textit{Following \textbf{Lemma 1} and \textbf{Lemma 4}, $\exists T_0\in\mathbb{N}$, s.t. when $t\ge T_0$, $|\delta_t|$ and the differences between elements converge, then we have: $\delta_t=0$ or $(\delta_t\ne 0)\land (|\delta_{t+1}/\delta_t|=1)$ (i.e, \textbf{\textcolor{blue}{(1)}} $d_t=0$ or \textbf{\textcolor{blue}{(2)}} $d_t=2\delta_t$). For \textbf{\textcolor{blue}{(1)}}, since $f(\cdot)$ is increasing and $q_t^i\in\{0,1,\cdots,K\}$, $\forall 1\le i\le l$, it holds that: $q_t^i-f(q_t^i-\delta_t)=f(q_t^i)-f(q_t^i-\delta_t)=0$, i.e, $\mathbf{Q}_t=\mathbf{Q}_{t+1}$. When $\delta_t>0$, $\exists q_t^i>0$ (otherwise $\delta_t\le 0$), so $[\text{round}(\delta_t)=0]\land[\text{frac}(\delta_t)\ne 1/2]$ or $[\text{floor}(\delta_t)=0]\land[\text{frac}(\delta_t)= 1/2]$ (see our \textbf{Proof} for \textbf{Lemma 2}), i.e., $0<\delta_t\le 1/2$. Similarly, when $\delta_t<0$, $\exists q_t^i<K$, so $\text{round}(-\delta_t)=0$, i.e., $-1/2<\delta_t<0$. For \textbf{\textcolor{blue}{(2)}}, using \textbf{Lemma 1}, it holds that: $q_t^i-f(q_t^i-\delta_t)=q_t^i-q_{t+1}^i=2\delta_t$. When $\delta_t>0$, using \textbf{Lemma 2} and \textbf{Lemma 3}, we have: $\text{round}(\delta_t)=2\delta_t$, i.e., $\delta_t=1/2$, but it leads $q_t^i-f(q_t^i-\delta_t)=0$ and is excluded. Similarly, when $\delta_t<0$, we have: $[-\text{round}(-\delta_t)]=2\delta_t$, i.e., $\delta_t=-1/2$, but it can also not hold since $q_t^i-q_{t+1}^i=-1$ itself will hold for limited steps. In summary, $\mathbf{Q}$ converges to a fixed $\mathbf{Q}_\infty$ as \textbf{\textcolor{blue}{(1)}} with a bounded $\delta_\infty\in(-1/2,1/2]$. $\blacksquare$}
 	
\begin{figure}
		\centering
		\hspace{-8pt}
		\vspace{-12pt}
		\includegraphics[width=0.495\textwidth]{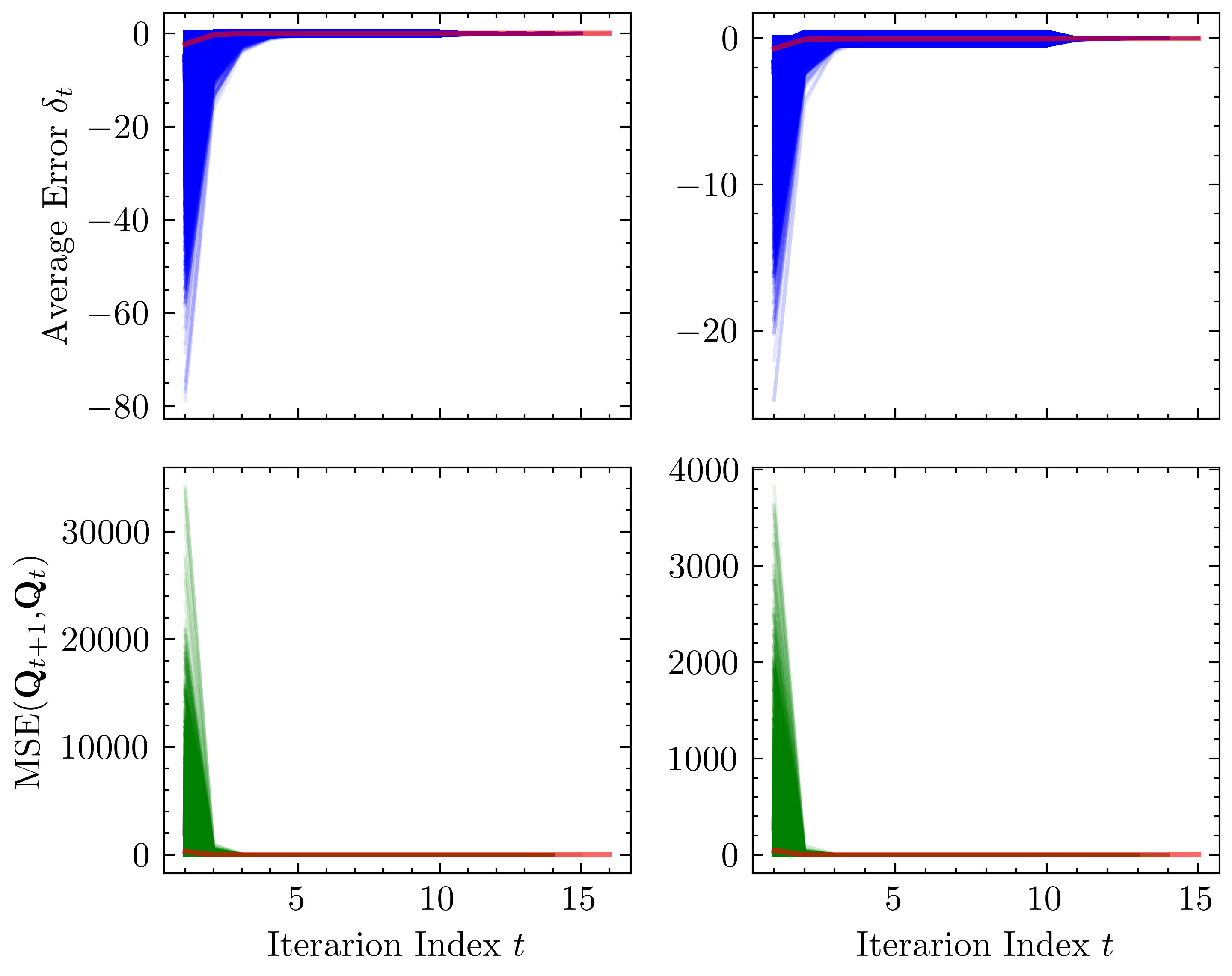}
		\vspace{-10pt}
		\caption{Visualization of the curves of average error $\delta_t$ (\textcolor{blue}{top}) and mean squared error of $\mathbf{Q}_{t+1}$ and $\mathbf{Q}_t$ (marked by $\text{MSE}(\mathbf{Q}_{t+1},\mathbf{Q}_{t})$, \textit{i.e.}, $({1}/{l})\lVert \mathbf{Q}_{t+1}-\mathbf{Q}_t \rVert _{F}^{2}$) (\textcolor{blue}{bottom}) under two settings of block number: $l=25$ (\textcolor{blue}{left}) and $l=100$ (\textcolor{blue}{right}) with the practical configurations $B=32$, $q=512$ and $K=N=1024$. The experiment consists of $1\times 10^5$ rounds for each setting with randomly generated initial states. Each blue or green curve corresponds to one our simulated BRA test round, and the red curve corresponds to the mean data values in each subgraph.}
		\label{fig: analysis_BRA}
		\vspace{12pt}
	\end{figure}

 	\vspace{4pt}
 	Second, after the preset limited $T$ uniform descent steps, the second part of BRA: random allocation (marked by \textcolor{blue}{method \#2}) will eliminate the remained error and converge since the sign of error is preserved and its absolute value monotonically decreases (proof is omitted). We'll get the accurately allocated measurement size map $\mathbf{Q}$ and CS ratio map $\mathbf{R}$ by our Alg. \ref{BRA}. Furthermore, Fig.~\ref{fig: analysis_BRA} visualizes the curves of average error and mean squared error of $\mathbf{Q}_{t+1}$ and $\mathbf{Q}_{t}$ under two settings, and demonstrates that our BRA strategy converges in 16 steps.
	
	\ifCLASSOPTIONcaptionsoff
	\newpage
	\fi

	
	
	%
	
	\bibliographystyle{IEEEtran}
	\bibliography{IEEEabrv,egbib}
	
\vspace{-1.2cm}
\begin{IEEEbiography}[\vspace{-0.8cm}{\includegraphics[width=1in,height=1.25in,clip,keepaspectratio]{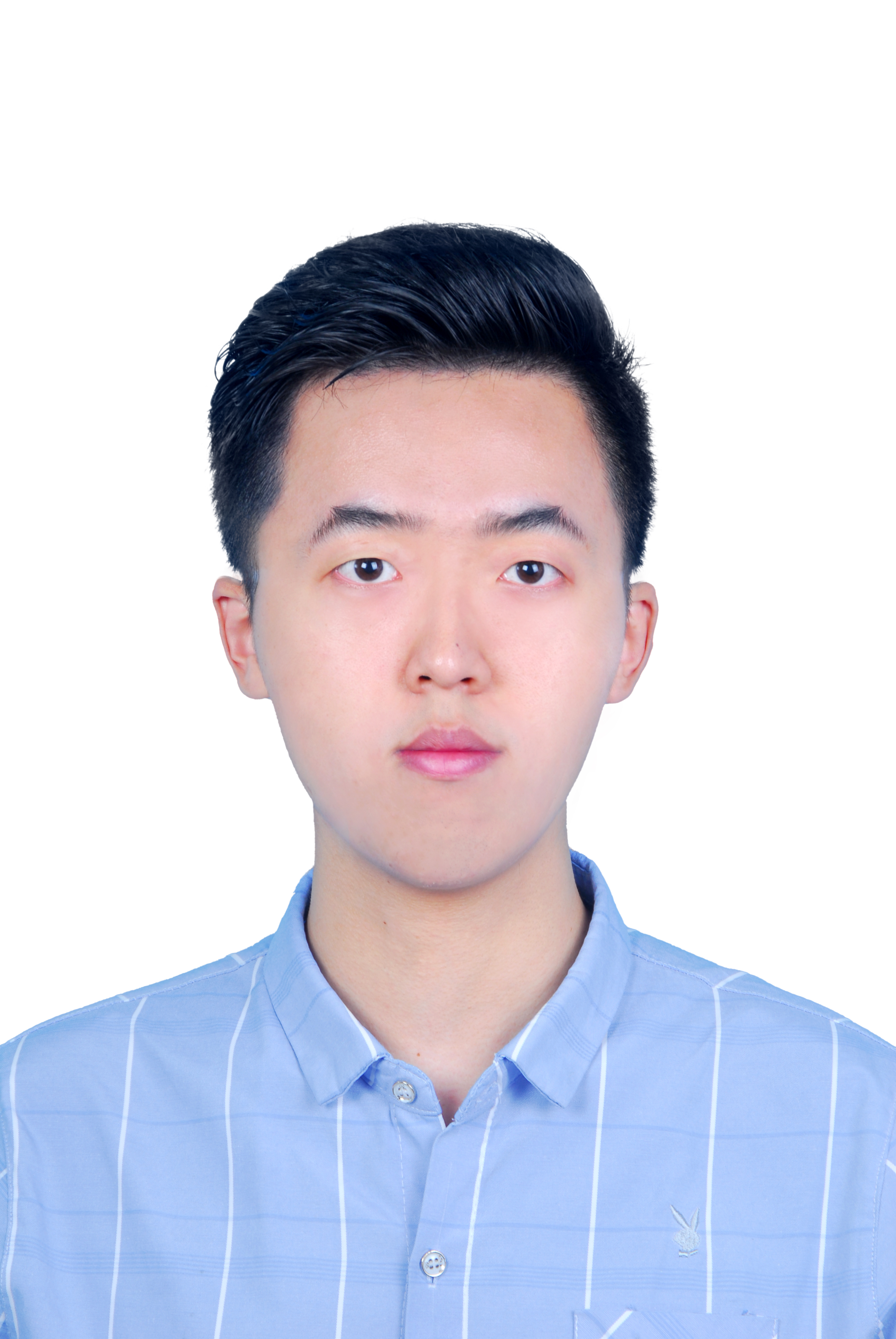}}]{Bin Chen} received the B.E. degree in the School of Computer Science, Beijing University of Posts and Telecommunications, Beijing, China, in 2021. He is currently working toward the master's degree in computer applications technology at Peking University Shenzhen Graduate School, Shenzhen, China. His research interests include compressive sensing, image restoration and computer vision. \end{IEEEbiography}

\vspace{-1.2cm}
\begin{IEEEbiography}[{\includegraphics[width=1in,height=1.25in,clip,keepaspectratio]{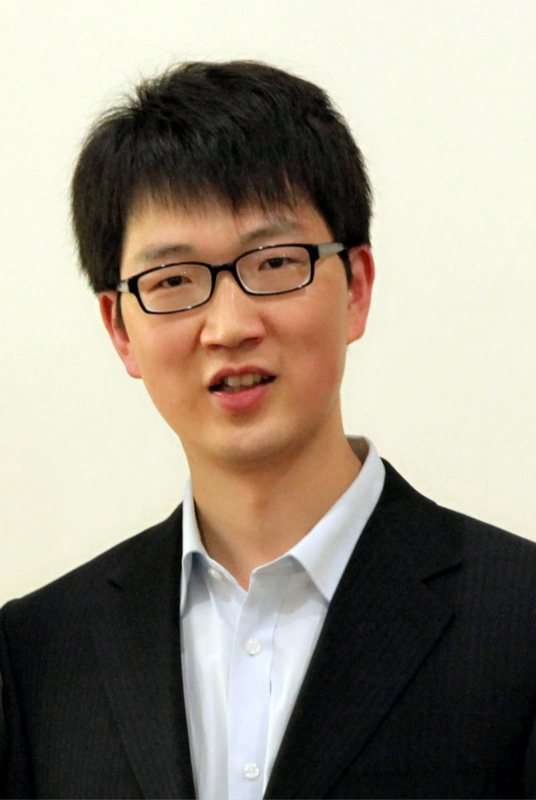}}]{Jian Zhang} (M'14) received the B.S. degree from the Department of Mathematics, Harbin Institute of Technology (HIT), Harbin, China, in 2007, and received his M.Eng. and Ph.D. degrees from the School of Computer Science and Technology, HIT, in 2009 and 2014, respectively. From 2014 to 2018, he worked as a postdoctoral researcher at Peking University (PKU), Hong Kong University of Science and Technology (HKUST), and King Abdullah University of Science and Technology (KAUST). 

Currently, he is an Assistant Professor with the School of Electronic and Computer Engineering, Peking University Shenzhen Graduate School, Shenzhen, China. His research interests include intelligent multimedia processing, deep learning and optimization. He has published over 90 technical articles in refereed international journals and proceedings. He received the Best Paper Award at the 2011 IEEE Visual Communications and Image Processing (VCIP) and was a co-recipient of the Best Paper Award of 2018 IEEE MultiMedia.
\end{IEEEbiography}
	
	%
	%
	
	%
	
	
	
	
	
	
	

\end{document}